%% file: main.tex
\documentclass[10pt]{article} 
\usepackage[preprint]{arxiv}

\usepackage{amssymb}            
\usepackage{mathtools}          
\usepackage{mathrsfs}           
\mathtoolsset{showonlyrefs}     
\usepackage{graphicx}           
\usepackage{subcaption}         
\usepackage[space]{grffile}     
\usepackage{url}                
\usepackage{booktabs}
\usepackage{xcolor}

\usepackage{algorithm}
\usepackage{algpseudocode}
\usepackage{wrapfig} 

\title{Heterogeneous Knowledge for Augmented Modular Reinforcement Learning}

\author{%
  Lorenz Wolf\\
  University College London\\
  \texttt{lorenz.wolf.22@ucl.ac.uk} \\
  \And
    Mirco Musolesi\\
    University College London\\
    University of Bologna\\
    \texttt{m.musolesi@ucl.ac.uk}\\
}

\begin{document} 
\pagestyle{fancy}
\fancyhead{}

\maketitle 
\begin{abstract}

Existing modular Reinforcement Learning (RL) architectures are generally based on reusable components, also allowing for ``plug-and-play'' integration.
However, these modules are homogeneous in nature - in fact, they essentially provide policies obtained via RL through the maximization of individual reward functions. Consequently, such solutions still lack the ability to integrate and process multiple types of information (i.e., heterogeneous knowledge representations), such as rules, sub-goals, and skills from various sources. In this paper, we discuss several practical examples of heterogeneous knowledge and propose Augmented Modular Reinforcement Learning (AMRL) to address these limitations. Our framework uses a selector to combine heterogeneous modules and seamlessly incorporate different types of knowledge representations and processing mechanisms. Our results demonstrate the performance and efficiency improvements, also in terms of generalization, that can be achieved by augmenting traditional modular RL with heterogeneous knowledge sources and processing mechanisms. Finally, we examine the safety, robustness, and interpretability issues stemming from the introduction of knowledge heterogeneity.
\end{abstract}


\input{preamble}

\newcommand{\BibTeX}{\rm B\kern-.05em{\sc i\kern-.025em b}\kern-.08em\TeX}


\section{Introduction}
Reinforcement learning (RL) is one of the most popular and successful approaches in machine learning, with its capability of optimizing decision-making processes in complex dynamic environments \citep{sutton_reinforcement_2015, kiran_deep_2021}. Its effectiveness is underscored by its application across diverse domains, such as robotics \citep{kober_reinforcement_nodate}, dynamic resource allocation \citep{waschneck_deep_2018}, and game playing \citep{mnih2015human, silver_mastering_2016, vinyals_alphastar_2019}.
Indeed, despite the remarkable progress made in RL, its practical applicability to real-world problems faces challenges related to sample efficiency and safety concerns \citep{garcia_comprehensive_2015}.
In fact, current model-free RL algorithms may require hundreds of million samples to converge to the optimal policy and exploration may cause the agent to take risky actions, which, in safety-critical situations, could lead to undesired outcomes \citep{garcia_comprehensive_2015}.

Modular RL is designed to tackle sample inefficiency by assembling multiple modules, each devoted to mastering specific "skills" for solving sub-problems; by integrating these modules, the system can address more intricate tasks collectively. This modular approach enhances efficiency by allowing the system to reuse learned skills across a variety of contexts, ultimately contributing to more effective problem-solving in complex environments
\citep{jacobs_adaptive_1991,russell_q-decomposition_2003, sprague_multiple-goal_2003, simpkins_composable_2019}. 
In general, modular approaches provide flexibility and can speed up learning when skills previously acquired can be transferred to solve new tasks \citep{devin_learning_2016}.
However, such solutions still lack the ability to process and integrate multiple types of information, such as rules, trajectory data, and skills. A collection of information from diverse knowledge sources with varying representations and processing mechanisms is what we call \textit{ heterogeneous knowledge}.

In many cases, existing heterogeneous knowledge could speed up the learning process and improve safety, however, current solutions are not flexible and generalizable. Indeed, efforts have been made to integrate different sources of information into RL, such as text manuals, video game-play training examples \citep{zhong_rtfm_2021, fan_minedojo_2022}, knowledge graphs \citep{murugesan_text-based_2021} and demonstrations \citep{abbeel_autonomous_2010, song_efficient_2012, garcia_safe_2012}. These methodologies are primarily designed to tackle specific issues and are constrained by their reliance on a \textit{single} source of information, thereby lacking the ability to incorporate \textit{diverse} types of knowledge representations.

\begin{figure*}
    \centering
    \includegraphics[width=0.815\textwidth, trim=0cm 4cm .0cm 0cm,clip]{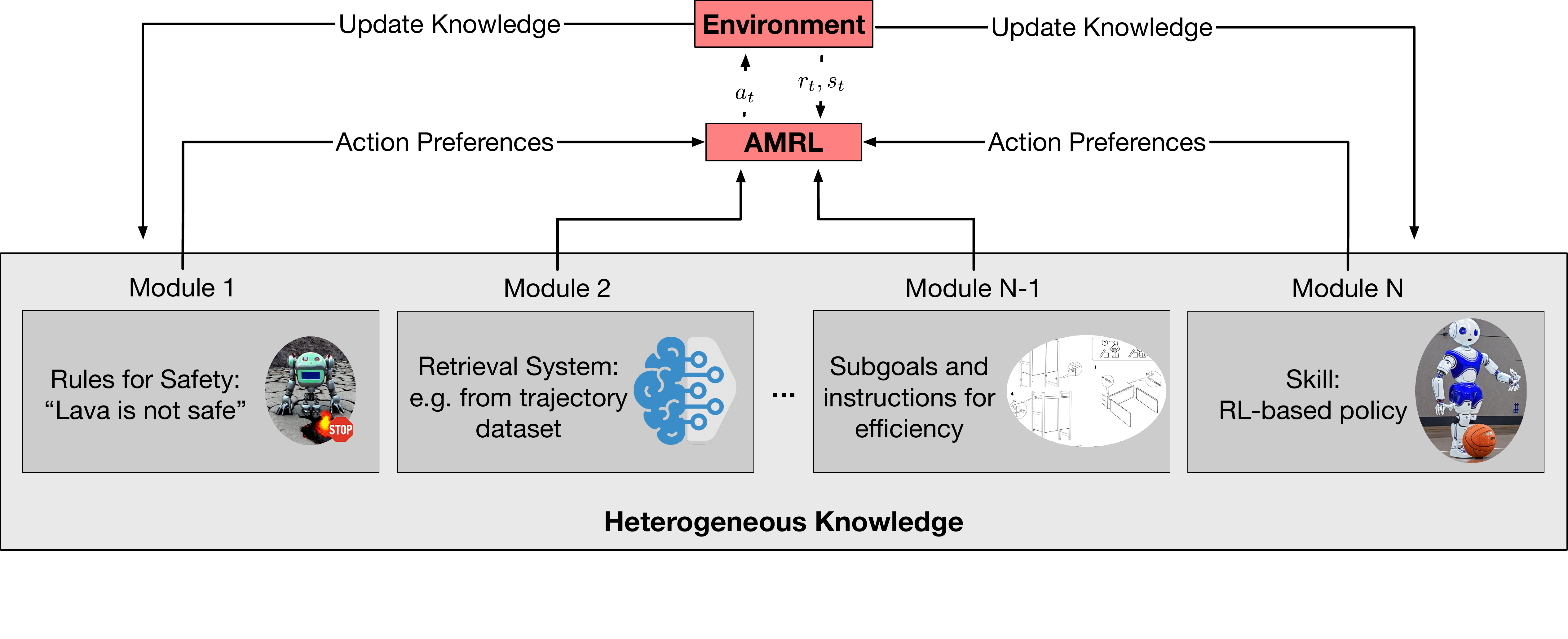}
    \caption{Examples of heterogeneous knowledge and the AMRL architecture. AMRL is able to access several sources of heterogeneous knowledge via modules. The modules can then be updated based on the environment feedback.}
    \label{fig:eyecatcher}
\end{figure*}

An agent that can seamlessly incorporate diverse knowledge sources and process them using a range of mechanisms is inherently compelling. As a motivating example consider an autonomous traffic control system at a busy intersection. Ideally, this system would: 1) Use existing rules based on logic (laws) and reason about them; 2) Utilize pre-trained policies trained on other intersections or a simulator; 3) Maximize traffic flow via standard RL; 4) Keep track of special cases in a database and retrieve scenarios when necessary, for example in an emergency situation. Consequently, this system not only relies on diverse knowledge representations, such as rules and past events but also processes these suitably. In our experiments we implement the four mentioned components for agents solving tasks of the Minigrid suite.

To instantiate such general agents, we propose Augmented Modular Reinforcement Learning (AMRL) (Figure \ref{fig:eyecatcher}). Building upon the framework of modular RL, we augment it through the seamless integration of various types of heterogeneous knowledge sources and processing mechanisms. Our approach not only includes modular skills but also incorporates rules, logic, and other forms of structured and/or declarative knowledge that can guide the agent's behavior more effectively. This augmentation not only strengthens the modularity of the RL system but also facilitates faster learning and ensures safer operations.
In our design, a general selector (or arbitrator) will have the role of selecting among heterogeneous modules. We investigate a hard selection mechanism corresponding to command arbitration \citep{simpkins_composable_2019}, which selects only one module to execute at a time, and a soft-selection mechanism resembling command fusion \citep{russell_q-decomposition_2003}, weighting the modules. Importantly, the selector itself is agnostic with respect to the actual knowledge representation. This allows for greater adaptability in diverse scenarios, making it suitable for a variety of applications\footnote{The code of AMRL is submitted in the supplementary materials and will be released publicly upon acceptance.}.
In this work we make the following two main contributions:
\begin{itemize}
    \item \textit{Formalization of a framework for the integration of heterogeneous modules for decision-making.} We introduce a formalism for designing heterogeneous modules for decision-making, enabling the integration of diverse types of knowledge representations. This provides a unified framework for expressing knowledge as rules, logic, skills (implemented as RL modules), and so on, which can be flexibly combined to tackle complex tasks. We discuss and implement several practical examples of heterogeneous modules.
    \item \textit{Augmentation of modular RL to incorporate heterogeneous knowledge.} We discuss the design of AMRL as an augmentation of a classical modular RL framework that enables the seamless integration of heterogeneous knowledge. Through this, we enhance the learning process, allowing the agent to leverage diverse prior knowledge for faster convergence. We compare two types of mechanisms, namely hard and soft selection. 
\end{itemize}
We evaluate the proposed architecture on a set of existing environments with different characteristics. We find that AMRL with the soft selection mechanism is more sample efficient than the baselines in all the environments and outperforms baselines on most of them (or it is able to achieve competitive performance). AMRL results in safer training and it is robust in presence of noisy modules. The hard selection mechanism results in noisier training behaviors impacting the overall learning process.

\section{Related work}

\paragraph{Knowledge augmented RL}
Grounding RL agents with external knowledge is an ongoing line of research in the field of RL. For example, progress has been made on incorporating knowledge provided to the agent via text for example as manual for a game \citep{zhong_rtfm_2021} or knowledge graphs connecting concepts and their characteristics \citep{murugesan_text-based_2021}. One approach to incorporate knowledge sources is the the formalism of a contextual MDP (CMDP) \citep{kirkSurveyZeroshotGeneralisation2023, perez2020generalized, ghosh2021generalization, hallak2015contextual}.
Another promising line of research focuses on equipping the agent with retrieval mechanisms to search over a memory of trajectories in order to inform the current decision \citep{goyal_retrieval-augmented_2022, humphreys_large-scale_2022}. However, knowledge is restricted to trajectories. To tackle open ended learning, \citet{LorangAdaptingTT} propose a hybrid hierarchical RL planning approach that uses skills for continuous robotic domains and \citet{rapidlearn} leverages symbolic knowledge to recover from fatal plan failures due to novelties.
To incorporate sub-optimal human knowledge \citet{zhang2020kogun} propose KoGuN. Related to KoGuN and most similar in motivation to our work are Knowledge Grounded Reinforcement Learning agents presented together with the KIAN architecture, which uses an attention mechanism to combine a learning agent with pre-trained policies~\citep{chiu_knowledge-grounded_2022, chiu2023flexible}. KIAN relies on rules as expert knowledge, while our focus is on the investigation and incorporation of heterogeneous knowledge.

\paragraph{Hierarchical and Modular RL}
Hierarchical RL (HRL) decomposes a problem into several smaller subtasks organized in a hierarchy \citep{digney_learning_nodate}. The higher-level parent-tasks can call lower-level child-tasks as if they were primitive actions. This enables higher-level tasks to focus on more abstract and longer-term learning while lower-level tasks are responsible for primitive actions and fine-grained control \citep{hengst_hierarchical_2010}. Each task is learned by a policy and at decision time; the higher-level policy triggers one in the level below. The goal is for the agent to perform task abstraction by focusing on more high-level tasks, so that it will suffer less from the curse of dimensionality \citep{barto_recent_2003}. This is a very active area of research with some of the earlier works dating back to the 1990s \citep{dayan_feudal_1992, vezhnevets_feudal_2017, parr_reinforcement_1997}.\\
HRL is closely related to modular RL, which aims to decompose the learning process into complementary policies/modules that can be combined by a controller \citep{sutton_horde_nodate, goyal_recurrent_2020, jacobs_adaptive_1991}. Previous works in Modular RL have considered command arbitration, selecting one module, which takes over the agent for the next action, and command fusion, combining the preferences of the different modules \citep{gupta_action_2021, russell_q-decomposition_2003, mendez_modular_2022, andreas_modular_2017}. We leverage the same modularity and flexibility for heterogeneous knowledge.

\paragraph{Command arbitration and fusion}
In the command arbitration paradigm an arbitrator decides which module takes control over the agent in a given state; the selected module then executes an action. Earlier works include GM-Sarsa \citep{sprague_multiple-goal_2003} and Q-decomposition \citep{russell_q-decomposition_2003}. More recent methods Arbi-Q \citep{simpkins_composable_2019} and GRACIAS \citep{gupta_action_2021} assume that the arbitrator receives a global reward signal to be maximized, enabling it to handle different scales of rewards for each of the modules. Furthermore, the global reward signal means that these methods can be used with off-policy as well as on-policy learning \citep{simpkins_composable_2019}.

\section{Preliminaries}
Let $\mathcal{M}=\left(\mathcal{S}, \mathcal{A}, \mathcal{P}, R, \rho_0, \gamma, T\right)$ represent a discrete-time finite- horizon discounted Markov decision process (MDP). $\mathcal{M}$ consists of the following elements: the state space $\mathcal{S}$, the action space $\mathcal{A}$, a transition probability distribution $\mathcal{P}: \mathcal{S} \times \mathcal{A} \times \mathcal{S} \rightarrow \mathbb{R}_{+}$, a reward function $R: \mathcal{S} \times \mathcal{A} \rightarrow \mathbb{R}$, an initial state distribution $\rho_0: \mathcal{S} \rightarrow \mathbb{R}_{+}$, a discount factor $\gamma \in[0,1]$, and the horizon $T$. Generally, the policy $\pi^{\theta}: \mathcal{S} \times \mathcal{A} \rightarrow \mathbb{R}_{+}$ is parameterized by $\theta$, which we optimize in order to maximize the expected discounted return under the policy. We denote the probability of taking action $a$ in state $s_t$ by $\pi^{\theta}(a|s_t)$ and abuse notation by denoting the distribution over all actions with $\pi^{\theta}(s_t)$.
Even if the architecture is conceptually independent from the underlying RL algorithm, the design and implementation presented in this paper relies on Proximal Policy Optimization (PPO) \citep{schulman2017proximal}.

\noindent Finally, in many real-world deployments, the underlying state is partially observable.
Formally, these scenarios can be modeled as a partially-observable Markov Decision Processes (POMDPs), which is represented by the tuple $\left(\mathcal{S}, \mathcal{A}, \mathcal{P}, \Omega, \mathcal{O}, R, \rho_0, \gamma, T\right)$, where in addition to the already known quantities we have introduced the observation space $\Omega$ and the observation probability distribution $\mathcal{O}(\cdot)$. Instead of $s\in \mathcal{S}$, the agent now observes $o\in \Omega$,  generated from the state $s$ via $o \sim \mathcal{O}(s)$.

\section{Heterogeneous Knowledge Sources, Representation, and Processing} 
\label{ssec:het_knowledge}

Heterogeneity can arise for various reasons. As a consequence, AMRL agents not only rely on \textit{heterogeneous knowledge} sources but also on {heterogeneous representation} and {processing mechanisms} (Table~\ref{tab:heterogeneous_knowledge}). Knowledge from the same source can be represented in various ways and, naturally, rules must be processed differently to trajectories.
We will adopt a bottom-up approach in the description of AMRL: we first discuss the modules at the basis of our approach; they essentially act as containers for heterogeneous knowledge sources and processing. In particular, we focus on the four types of modules outlined in the motivating example, specifically, rules, skills, retrieval, and dynamic RL. We then discuss how these modules are incorporated in the overall architecture via a selection mechanism based on hierarchical modular RL.
\begin{table}[t]
    \centering
    \small
    \caption{Different types of heterogeneous knowledge.}
    \label{tab:heterogeneous_knowledge}
    
    \begin{tabular}{ccc}
        \hline
        \textbf{Source} & \textbf{Representation} & \textbf{Processing}\\
        \midrule
        Human Expert & Rules & Logic\\
        Environment & RL-based policy & Skill Execution\\
        Previous Deployment & Trajectory Database & Retrieval\\
        \hline
    \end{tabular}
\end{table}
\subsection{Knowledge Representation and Processing using Modules}
\label{ssec:knowledge}
An AMRL agent is instantiated with several modules $M_i \in \mathbf{M}$, which each act as a container for knowledge representation $K_i$ and a corresponding processing mechanism. Module $M_i$ returns action preferences, i.e., probabilities over actions, based on its knowledge source giving $\pi_{M_i}(a|s_t) = M_i(a,s_t|K_i)$. In contrast to previous works in modular RL, the AMRL modules are heterogeneous. They are not limited to RL policies and can be diverse in terms of how they form preferences over actions. Modules and knowledge can be static or dynamic. If the knowledge source and, consequently, the mapping of a module $M_i$ is to be updated during training, modular feedback must be observed, such that the effect of actions proposed by $M_i$ can be learned.
Modular rewards $\{r_{i,t}\}_{i=1}^N$, where $r_{i,t}$ is the reward for module $M_i$ at time $t$, can be specified to favor modular learning. In case $M_i$ is static no modular updates are performed.

We now focus on the knowledge $K_i$ contained in modules $M_i$. Knowledge sources can be represented in many different forms, including rules, manuals, trajectories, etc. and can be assessed according to the following characteristics: \textit{actionability}, \textit{interpretability}, and \textit{informativeness}. A knowledge representation is \textit{actionable} if it can be used directly to inform the action selection process.
Most actionable choices are procedural or declarative knowledge sources, for example in form of a world model \citep{hafner_mastering_2023} or a database of past trajectories \citep{goyal_retrieval-augmented_2022, humphreys_large-scale_2022}.
While these might be the most effective representations for RL agents, they are less interpretable and importantly can be harder to obtain in the first place due to deployment restrictions. Other more abstract knowledge representations such as rules and manuals are more \textit{interpretable}, easier to validate and in many cases more widely available, which highlights the importance of flexibly incorporating and acquiring heterogeneous knowledge. The \textit{informativeness} captures how relevant and useful the knowledge is for solving the given task. In our evaluation we compare the effect of these properties on the agent's performance and sample efficiency.

\subsection{Types of Modules}
\label{ssec:modules}
In this subsection we provide practical instantiations of heterogeneous modules. It is worth noting that the proposed architecture allows for other types of modules based on different knowledge representations and processing mechanisms. We select modules that are representative of broader classes of solutions. We are also aware of ongoing research on each of these but note that our focus is on combining heterogeneous decision-making mechanisms. Deeper investigation into each possible mechanism is beyond the scope of this paper.

\subsubsection{Logic-based rules.}
This module relies on a set of rules $\Lambda=\{\lambda_i\}_{i=1}^{N_\lambda}$, where each rule $\lambda_i$ is formalized as statement of the form: $$\text{ If } s_t \text{ satisfies condition } c \text{ then } \pi_M(a|s_t) = p_{\lambda_i}(a),$$ where $p_{\lambda_i}(a)$ is the probability of taking action $a$ under rule $\lambda_i$. For example, let us consider one of the typical benchmark games where an agent has to move in a space where a lava flow is present: if there is lava in front of the agent then the probability of moving forward is set to $0$.
Conversely, let us suppose to have a different game environment in which, for instance, an agent has to collect objects, such as keys: if there is a key in front of the agent then the probability of picking up the key is set to $1$. Rules can be ordered in a hierarchy and one rule may call on other rules. Furthermore, rules can contradict or coincide in which case the conflict needs to be resolved. As this is not the main focus of this work, we simply resolve such conflicts by averaging the probabilities over the set of rules applicable in state $s_t$ denoted by $\Lambda_t$, such that: $$ \pi_M(a|s_t, \Lambda) = \frac{1}{|\Lambda_t|} \sum_{\lambda\in \Lambda_t} p_\lambda(a).$$

\subsubsection{Trajectory database with retrieval.}
This module relies on a database containing trajectories and retrieves relevant information via nearest-neighbor search. The knowledge source in this case is given by the set $\mathcal{D}=\{\tau_i\}_{i=1}^{N_{D}}$, where each $\tau_i = ((s_{i,1}, a_{i,1}, r_{i,1}), \dots,$ $(s_{i,T_i}, a_{i,T_i}, r_{i,T_i}))$ is a trajectory of length $T_i$ containing state-action-reward tuples. The retrieval mechanism relies on an embedding network mapping from the state space to an embedding space. Given the current state the query is formed via $q_t = embed(s_t)$. The $k$ approximate nearest-neighbors in the embedding space are retrieved from $\mathcal{D}$ with FAISS \citep{johnson2017billionscale} based on the $L_2$-norm, which yields the k tuples $(s_{n_1}, a_{n_1}, r_{n_1}),\dots,(s_{n_k}, a_{n_k}, r_{n_k})$ with corresponding distances $d_{n_1}, \dots, d_{n_k}$ to the query. The retrieved information is utilized to form an action by weighting according to rewards and distance, which yields action preferences:
$$ \pi_M(a|s_t, \mathcal{D}) = \frac{e^{p_D(a)}}{\sum_{a\in\actionspace}e^{p_D(a)}}, \text{ with } p_D(a) = \sum_{j=1}^k  \frac{r_{n_j}}{d_{n_j}} I(a=a_{n_j}).$$
Other more sophisticated methods for retrieval such as those proposed by \citet{goyal_retrieval-augmented_2022} and \citet{humphreys_large-scale_2022} can be substituted instead.

\subsubsection{RL-based policies (skills).}
The third class of modules we consider is that based on skills, i.e. on policies $\pi_{skill}$ trained via RL with an unknown reward function. The policies are trained on the same state space $\statespace$ with action space $\actionspace_{skill} \subset \actionspace$. These modules simply contain the policy and return corresponding action probabilities given by $\pi_M(a|s_t)= \pi_{skill}(a|s_t)$ if $a\in \actionspace_k$ and $0$, otherwise.

\section{Combining Heterogeneous Knowledge} 
To incorporate various types of knowledge sources and processing mechanisms as described in Section \ref{ssec:het_knowledge}, we employ a controller / arbitrator, which we refer to as the \textit{selector}. The goal of this component is to select/combine one or more modules. In particular, it determines which modules to trigger in a given state, thereby enabling the performance of tasks with varying levels of complexity.

We consider two variants of selection mechanisms, which correspond to command arbitration and command fusion. The first, which we refer to as \textit{hard selection}, restricts the weights to a one-hot vector so that exactly one module is selected and executed at each time step. The second, which we refer to as \textit{soft selection}, relies on an attention mechanism to form a weighted average combining preferences from several modules, resulting in command fusion. 
Thus, by weighting the modules, this variant of the selector is based in a sense on a mixture model.
We denote the selector's policy mapping from the state space to the selector's action space by $\pi^{\phi}_{selector}: \statespace \rightarrow\actionspace_{selector}$,  parameterized by $\phi$. We have $\actionspace_{selector} = \{M_1, \dots, M_N\}$, such that $\pi^{\phi}_{selector}(M_i|s_t)$ is the probability of choosing module $M_i$ in state $s_t$ and expresses the selector's preference for module $M_i$. In the following, we will provide a formalization of the two selection mechanisms and discuss how to bridge the gap between the current task and what is possible by flexibly combining the modules.

\subsubsection{Hard selection.}
At each time step $t$, the selector observes the current state $s_t$ and selects exactly one module $M$ to execute. We then sample an action from the chosen module $a_t\sim \pi_M(s_t)$. In particular, one module is sampled according to the selector's policy $M \sim \pi^{\phi}_{selector}(s_t)$. To perform hard selection differentiably, we obtain a sample vector $\mathbf{y}$ of dimensions $|\actionspace_{selector}|$ by setting:
$$y_i=\frac{\exp \left(\left(\log \left(\pi^{\phi}_{selector}(M_i|s)\right)+g_i\right) / \tau\right)}{\sum_{j=1}^N \exp \left(\left(\log \left(\pi^{\phi}_{selector}(M_j|s)\right)+g_j\right) / \tau\right)}$$
\noindent for $i=1, \ldots, N$,where $g_1 \ldots g_N$ are i.i.d samples drawn from\\ $\operatorname{Gumbel}(0,1)$, which can be sampled using an inverse transform \citep{jang2017categorical}. The obtained sample $\mathbf{y}$ is subsequently discretized into a one-hot vector. This ensures the selection of exactly one module at a time. A soft approximation is used for the computation of the gradients. This yields as the AMRL agent's policy $\pi(a|s_t) = \pi_M(a|s_t)$ with $M\sim \pi^{\phi}_{selector}(s_t)$. In the experiments we set the temperature $\tau=1$. We provide an ablation analysis in Appendix \ref{app:gumbel}.

\subsubsection{Soft selection.}
In contrast to hard selection, soft selection combines the modules' policies with a weighted average. The action preferences $\pi_{M_i}(a|s_t)$ returned by the modules are weighted by the module preferences $\pi^{\phi}_{selector}(M_i|s_t)$ returned by the selector. In particular, we have that $\pi(a|s_t) = \sum_{i=1}^N \pi^{\phi}_{selector}(M_i|s_t)\pi_{M_i}(a|s_t)$. Note that $\pi^{\phi}_{selector}(s_t)$ is normalized to sum to $1$.

\subsubsection{Implementing the selectors.} In many applications, flexibly combining the knowledge available to the agent via the modules may not be sufficient to achieve high performance on a new task. Similarly to the inner actor in the KIAN architecture, we bridge this gap by adding a dynamic RL module. This module contains a learnable RL policy $\pi^{\theta}_{M_{dyn}}: \statespace \times \actionspace \rightarrow \mathbb{R}_+$ trained to maximize the global reward function also observed by the selector.
In practice, the selector is implemented as a neural network parameterized by parameters $\phi$, which, given the current state as input, outputs probabilities over the modules. By including the dynamic RL module, we have as AMRL policy with soft selection:
\begin{align}
    \pi^{\phi,\theta}(a|s) = &\pi_{selector}^{\phi}(M_{dyn}|s)\pi_{M_{dyn}}^{\theta}(a|s) + \sum_{M\in \mathbf{M}\setminus M_{dyn}} \pi_{selector}^{\phi}(M|s)\pi_{M}(a|s),\label{eq:softpolicy}
\end{align}
\noindent which allows us to easily compute policy gradients w.r.t. $\phi$ and $\theta$, justifying the use of PPO to train the selector (see Appendix \ref{app:gradients} for details). Note again that the selector itself is agnostic with respect to the actual knowledge representation. The entire decision-making process is presented in Algorithm~\ref{alg:selector}.

\subsubsection{Module prioritization.}
With heterogeneity come inherent differences between the modules. In particular modules such as the rule module may be considered safe and trustworthy, carefully designed and well understood. Consequently, while the rules may not be applicable in all cases, if applicable, they can be trusted and should be followed by the agent. This can be achieved with module prioritization. In  the case of the rule module, whenever one or more of its rules are triggered, the rules act as a constraint, i.e., other modules' action preferences can be taken into account, but only as long as the agent does not violate any of the triggered rules. This can also be seen as masking of actions. In other words, for a given state $s$, the actions, which are not to be taken, are extracted from e.g. the rule module in form of a mask defined by $\Lambda_a = 0$ if $a$ is an unsafe action and $1$ otherwise. Then the final policy is given by $\Lambda_a \pi^{\phi,\theta}(a|s)$ where $\pi^{\phi,\theta}(a|s)$ is defined in Equation \eqref{eq:softpolicy}. In safety critical states, the applicable rules mask unsafe actions, which are not to be taken. This reduces the percentage of unsafe actions to $0$ given that the rules cover all cases. Similarly to the soft and hard selection mechanisms, a soft prioritization mechanism based on specified weights can also be implemented. 
\begin{algorithm}[t]
\caption{Decision-making with an AMRL agent.}\label{alg:selector}
\begin{algorithmic}
\Require Modules and knowledge $\mathbf{M} = \{M_{i}, K_i\}_{i=1}^N \cup M_{dyn}$

\State $s_0 \gets \text{Initial state}$
\For{each timestep $t=0,...,T$}
    \State Get module preferences $\pi_{M_i}(s_t) \gets M_{i}(s_t|K_i)$
    \If{Hard selection}
        \State Sample selected module $M \sim \pi^{\phi}_{selector}(s_t)$
        \State Get AMRL policy $\pi(a|s_t) = \pi_M(a|s_t)$
    \ElsIf{Soft selection}
        \State $\pi(a|s_t) = \sum_{M\in \mathbf{M}} \pi^{\phi}_{selector}(M|s_t)\pi_{M}(a|s_t)$
    \EndIf
    \State Sample action from AMRL policy $a_{t} \sim \pi(s_t)$
    \State Observe $s_{t+1}$ and reward $r_{t}$
    \State Update selector parameters $\phi$ and dynamic module parameters $\theta$ with PPO
\EndFor

\end{algorithmic}
\end{algorithm}

\begin{figure*}[t]
    \centering
    \begin{subfigure}[b]{.24\textwidth}
        \centering
        \includegraphics[width=\textwidth]{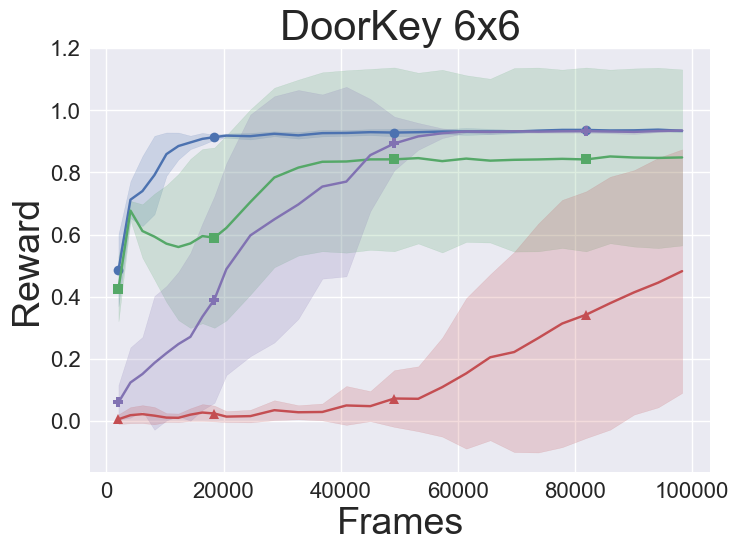}
    \end{subfigure}
    \hfill
    \begin{subfigure}[b]{.24\textwidth}
    \centering
        \includegraphics[width=\textwidth]{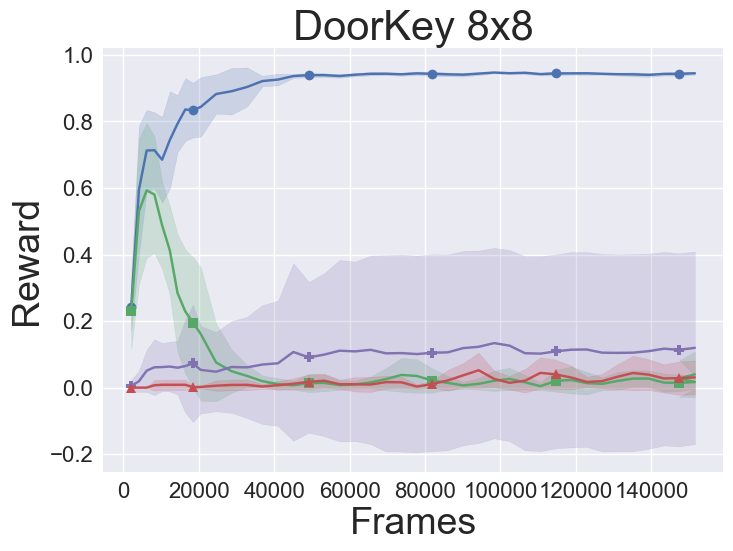}
    \end{subfigure}
    \hfill
    \begin{subfigure}[b]{.24\textwidth}
    \centering
        \includegraphics[width=\textwidth]{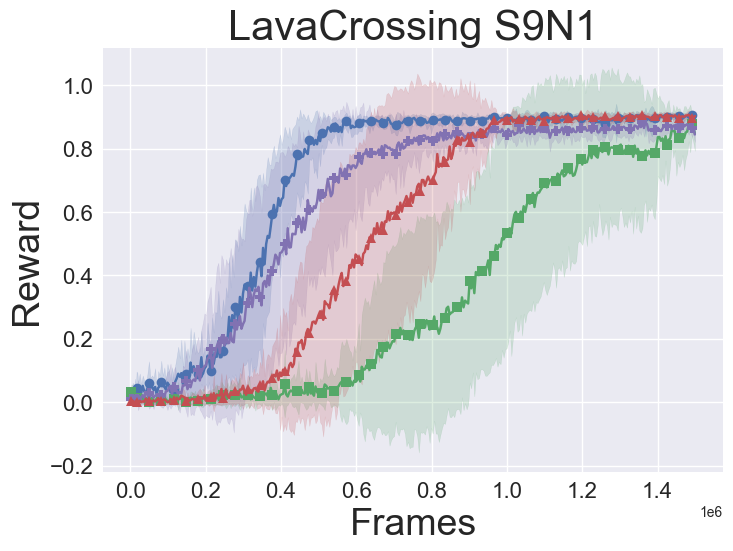}
    \end{subfigure}
    \hfill
    \begin{subfigure}[b]{.24\textwidth}
    \centering
        \includegraphics[width=\textwidth]{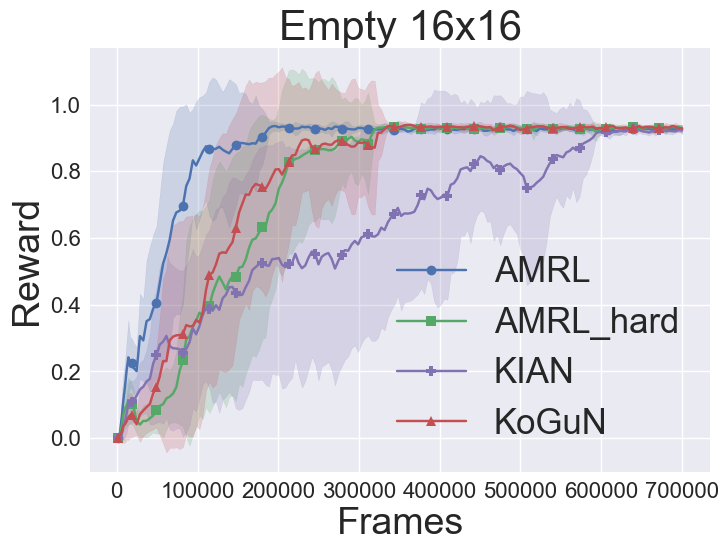}
    \end{subfigure}
    \caption{The achieved reward logged throughout training. On all 4 environments AMRL with soft selection uses heterogeneous knowledge to achieve good performance more efficiently than baselines. The hard selection mechanism strongly limits its capabilities and results in significantly nosier behaviors.}
    \label{fig:env_gen_perf}
\end{figure*}

\section{Evaluation}
In the evaluation of AMRL (and solutions for the integration of heterogeneous knowledge in RL more generally) we focus on the overall performance and sample efficiency, the dependence on knowledge informativeness and robustness against random modules, the safety gains during training and at test time, and the interpretability of the selection mechanism.

 \subsection{Experimental Settings and Baselines}

\subsubsection{Overview.} We now evaluate AMRL considering a range of modules by means of different benchmark environments. In particular, we train the agents on several environments contained in the Minigrid suite \citep{minigridenv}. These are characterized by different levels of complexity and include safety-critical ones such as Lava-Crossing. We provide a more detailed description of these environments in Appendix~\ref{app:envs}. Unless specified otherwise explicitly, we use the acronym AMRL to denote AMRL with the soft selection mechanism. 
We consider instances of the module types described in Section~\ref{ssec:modules}, which also correspond to the modules outlined in our motivating example. The retrieval module relies on a dataset of trajectories collected from an expert on a specified task. The skill module is a pre-trained policy from a specified environment, and the rules module contains a set of 6 rules relevant to the Minigrid environments, such as ``do not step onto lava''. If that rule is triggered, the module will assign probability zero to the action that would lead to the agent stepping on lava (more details in Appendix~\ref{app:rules}). All experiments are performed with 10 random seeds and we report mean and sample standard deviation.\\
\subsubsection{Baselines.}
As baselines, we implement \textbf{KIAN} \citep{chiu2023flexible}, \textbf{KoGuN} \citep{zhang2020kogun}, and standard \textbf{PPO} \citep{schulman2017proximal}. KIAN and KoGuN are two cutting-edge methods for knowledge augmented RL and are adapted to our scenario in which we deal with heterogeneous knowledge to make them as competitive as possible. The implementations of all agents and PPO rely on the \verb+rl-starter-files+ repository and \verb+torch_ac+\footnote{The code of the libraries can be found at the following URLs: \href{https://github.com/lcswillems/rl-starter-files}{https://github.com/lcswillems/rl-starter-files} and \href{https://github.com/lcswillems/torch-ac}{https://github.com/lcswillems/torch-ac}}. We use the default hyperparameter settings for PPO provided in \verb+rl-starter-files+. Additional implementation details are provided in Appendix~\ref{appendix:reproducibility}.
\textbf{KIAN} learns a key for each piece of knowledge (originally rules) via an embedding layer. The actor consists of an inner component, which plays the same role as the dynamic RL module in our architecture, a query network that, given the current state, forms a query used to weight the knowledge pieces, and another fully connected layer to learn the keys of each knowledge piece. 
We have directly adopted the official implementation. 
\textbf{KoGuN} implemented with heterogeneous knowledge corresponds to evaluating each module, and concatenating the outputs to $k(s_t) = (\pi_{M_1}(s_t),...,\pi_{M_K}(s_t))$. The knowledge vector $k(s_t)$ is then fed into the actor network as part of the state representation, such that $\pi(a|s_t) = \pi_{\theta}(k(s_t), s_t)$. 
\textbf{PPO} is our baseline that does not incorporate any additional knowledge.

\subsubsection{Computational Cost}
The computational cost for training and deploying AMRL is comparable with that of KoGuN. KIAN has a larger number of trainable parameters and a slightly slower inference time. PPO is faster at inference time since it does not rely on any of the knowledge modules. The details of this analysis can be found in Appendix \ref{app:cost}.

\subsection{Heterogeneous Knowledge for Efficiency Across Environments}
To evaluate how heterogeneous knowledge improves learning efficiency across a range of environments all agents are equipped with the following knowledge. In particular, we consider \textit{Skill}, which solves the Minigrid Unlock environment; \textit{Retrieval}, which is based on trajectory data collected in Empty Random 5x5 by an agent trained on the Empty Random 5x5 task; and, finally, \textit{Rules},  which relies on all $6$ rules for Minigrid (details in Appendix \ref{app:rules}).
AMRL and baselines benefit from heterogeneous knowledge. In particular, AMRL with soft selection is more efficient and achieves higher performance than baselines across most environments (see Table~\ref{tab:perf_envgen}). Not only does AMRL achieve strong final performance, but Figure~\ref{fig:env_gen_perf} also shows its improved sample efficiency across several Minigrid environments. Notably, the performance difference is largest on the DoorKey 8x8 environment for which other methods have not managed to make meaningful progress by the time AMRL has solved the environment. 

While AMRL with soft selection performs well, the hard selection mechanism leads to ``noisy'' behavior and poorer performance, often with slower convergence and without reaching competitive final performance. This should be expected due to its inability of combining the action preferences from several modules, but it is also dependent on the temperature parameter $\tau$ (see Appendix \ref{app:gumbel}).
We note that all three modules are highly informative for solving the DoorKey environments in contrast to only 2 modules for Empty (Retrieval \& Rules) and only 1 module for Lava Crossing (Rules). This is apparent observing the relative performance of agents accessing knowledge against PPO, which improves as the module informativeness decreases.

\begin{figure}[t]
    \centering
    \begin{subfigure}{0.45\textwidth}
    \centering
    \includegraphics[width=\linewidth, trim={0 0mm 0mm 5mm}, clip]{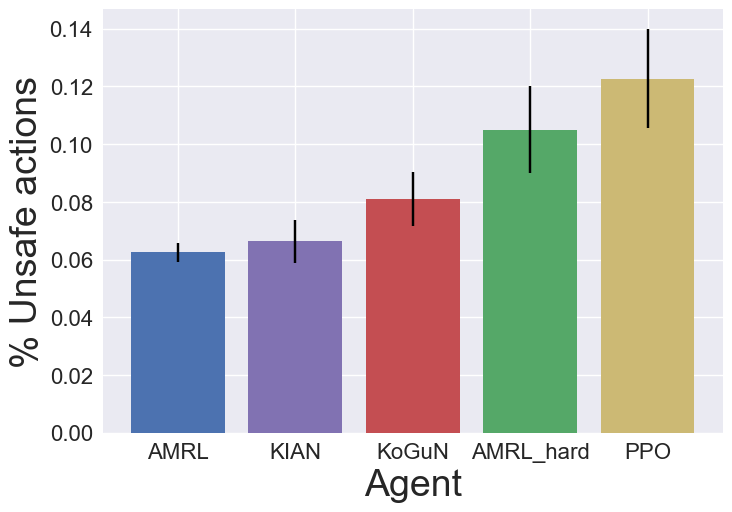}
    \caption{}
    \label{fig:safety}
    \end{subfigure}
    \hfill
    \begin{subfigure}{0.45\textwidth}
    \centering
    \includegraphics[width=\textwidth, trim={0 0mm 0mm 5mm}, clip]{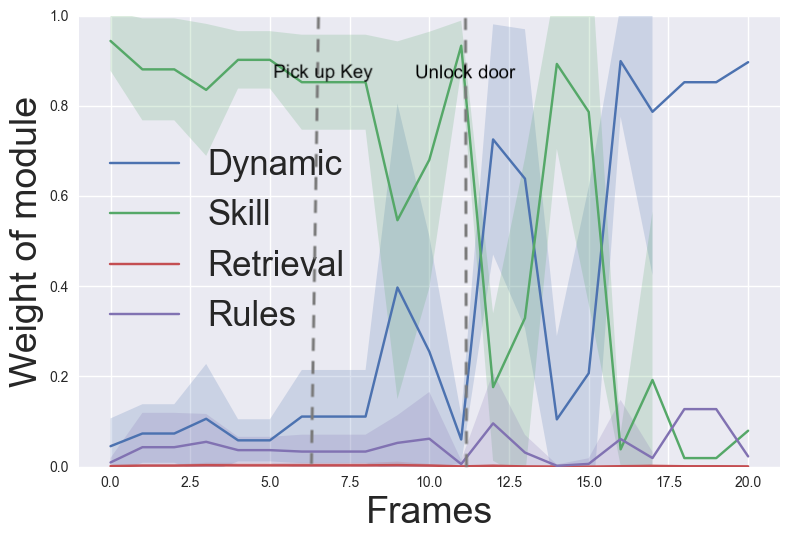}
    \caption{}
    \label{fig:selector:weights}
    \end{subfigure}
    \caption{\ref{fig:safety}) Percentage of unsafe actions during training on LavaCrossing S9N1. Confidence intervals are $\pm$ 2 standard deviations across 10 random seeds. AMRL performs the smallest amount of unsafe actions closely followed by KIAN. \ref{fig:selector:weights}) AMRL selector weights during one episode evaluation on DoorKey 8x8 showing mean $\pm 2$ standard deviation calculated across $10$ training seeds. On average, agents pick up the key with their 6th action and unlock the door with their 11th action, indicated by the dashed lines. The second milestone causes a shift in the module weights away from the unlock skill and towards the dynamic module.}
\end{figure}

\begin{table*}[t]
    \centering
    \caption{Evaluation of heterogeneous knowledge for generalization and efficiency across environments. Agents use the following heterogeneous knowledge: \textit{skill=''Unlock"}, \textit{retrieval$=$"Empty\_random\_5"}, \textit{rules$=$"all"}. AMRL with soft selection outperforms except for the Lava S9N2 environment (note large standard deviations).}
    \label{tab:perf_envgen}
    \begin{tabular}{lccccc}
        \hline
          & DoorKey 6x6 & DoorKey 8x8 & Lava S9N1 & Lava S9N2 & Empty 16x16 \\
        \midrule
        AMRL & \textbf{0.93} (0.005) & \textbf{0.94} (0.005) & \textbf{0.90} (0.009) & 0.54 (0.412) & \textbf{0.93} (0.009) \\
        AMRL$_{hard}$ & 0.84 (0.285) & 0.02 (0.027) & 0.86 (0.066) & 0.02 (0.036) & \textbf{0.93} (0.010) \\
        KIAN & \textbf{0.93} (0.005) & 0.11 (0.293) & 0.87 (0.021) & \textbf{0.62} (0.237) & \textbf{0.92} (0.007) \\
        KoGuN & 0.32 (0.359) & 0.03 (0.040) & \textbf{0.90} (0.011) & 0.52 (0.441) & \textbf{0.93} (0.009) \\
        PPO & 0.13 (0.200) & 0.01 (0.018) & 0.62 (0.357) & 0.07 (0.156) & \textbf{0.93} (0.011) \\
        \hline
    \end{tabular}
    
\end{table*}
\begin{table*}[th]
    \centering
    \caption{Mean reward (one standard deviation) with different knowledge of varying informativeness levels. Agents on DoorKey 8x8 are trained for 300k frames and on Empty 16x16 for 200k frames. The details of knowledge modules are given in Appendix \ref{app:exp_details}. AMRL consistently benefits from more informative modules, particularly achieving near-optimal performance with highly informative knowledge in both environments. In contrast, AMRL$_{hard}$, KIAN, and KoGuN exhibit limited improvements with increasing access to informative knowledge, suggesting that they struggle to fully leverage available knowledge.}
    \label{tab:informativeness}
    \begin{tabular}{lcccccc}
    \hline
     & \multicolumn{3}{c}{DoorKey 8x8} & \multicolumn{3}{c}{Empty 16x16} \\
     & low & medium & high & low & medium & high \\
    \midrule
    AMRL & 0.03 (0.03) & \textbf{0.12} (0.28) & \textbf{0.93} (0.01) & \textbf{0.86} (0.23) & \textbf{0.84} (0.22) & \textbf{0.88} (0.18) \\
    AMRL$_{hard}$ & 0.02 (0.02) & 0.03 (0.03) & 0.07 (0.06) & 0.65 (0.3) & 0.71 (0.31) & 0.65 (0.31) \\
    KIAN & 0.03 (0.04) & 0.04 (0.06) & 0.27 (0.39) & 0.61 (0.35) & 0.58 (0.38) & 0.63 (0.34) \\
    KoGuN & 0.02 (0.03) & 0.03 (0.028) & 0.16 (0.31) & 0.73 (0.3) & 0.79 (0.3) & 0.79 (0.27) \\
    \hline
    \end{tabular}   
\end{table*}

\begin{table*}[th]
    
    \centering
    \caption{Evaluating robustness in presence of random modules. Agents on LavaCrossing S9N1 for 1.5 million frames, on Empty 16x16 for 150k frames, DoorKey 8x8 for 300k frames, DoorKey 6x6 for 200k frames. We report final reward $\pm 2$ standard deviations across 10 random training seeds. AMRL is the most robust, closely followed by KoGuN. KIAN shows moderate robustness, and AMRL with hard selection is the most sensitive to random modules.}
    \label{tab:noisy}
    \tiny
    \begin{tabular}{lccccccccc}
    \hline
     & \multicolumn{4}{c}{Lava S9N1} & \multicolumn{4}{c}{Empty 16x16} \\
    \#random & 0 & 1 & 3 & 5 & 0 & 1 & 3 & 5\\
    \midrule
    AMRL & 0.90 (0.009) & 0.89 (0.014) & 0.90 (0.011) & 0.90 (0.012) & 
    0.92 (0.053) & 0.78 (0.264) & 0.73 (0.290) & 0.90 (0.077) \\
    AMRL$_{hard}$ & 0.86 (0.066) &  0.86 (0.067) & 0.21 (0.392) & 0.01 (0.013) &  
    0.60 (0.390) & 0.22 (0.267) & 0.14 (0.108) & 0.06 (0.086)\\
    KIAN & 0.87 (0.021) &  0.85 (0.026) & 0.79 (0.064) & 0.75 (0.057) & 
    0.41 (0.220) & 0.31 (0.234) & 0.25 (0.077) & 0.25 (0.104)\\
    KoGuN & 0.90 (0.011) &  0.90 (0.009) & 0.90 (0.016) & 0.89 (0.025) & 
     0.50 (0.382) & 0.63 (0.341) & 0.46 (0.303) & 0.45 (0.281) \\
    \hline
    & \multicolumn{4}{c}{DoorKey 8x8} & \multicolumn{4}{c}{DoorKey 6x6} \\
    \#random & 0 & 1 & 3 & 5 & 0 & 1 & 3 & 5\\
    \midrule
    AMRL & 0.94 (0.005) & 0.95 (0.003) & 0.57 (0.510) & 0.70 (0.404) & 
    0.94 (0.003) & 0.94 (0.004) & 0.94 (0.004) & 0.94 (0.006) \\
    AMRL$_{hard}$ & 0.04 (0.029) & 0.03 (0.038) & 0.01 (0.014) & 0.02 (0.026) &  0.94 (0.005) & 0.75 (0.417) & 0.44 (0.459) & 0.02 (0.030) \\
    KIAN & 0.61 (0.444) & 0.05 (0.069) & 0.26 (0.444) & 0.26 (0.376)  & 
    0.93 (0.004) & 0.93 (0.004) & 0.92 (0.006) & 0.92 (0.005)\\
    KoGuN & 0.22 (0.405) & 0.07 (0.040) & 0.04 (0.032) & 0.02 (0.024) & 
    0.93 (0.006) & 0.93 (0.005) & 0.93 (0.005) & 0.93 (0.006) \\
    \hline
    \end{tabular}   
\end{table*}
\begin{figure*}[th]
    \centering
    \begin{subfigure}[b]{.24\textwidth}
        \centering
        \includegraphics[width=\textwidth]{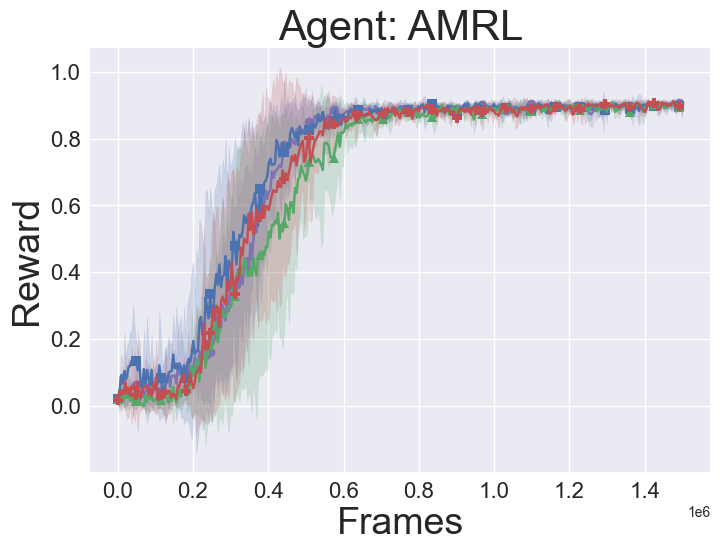}
    \end{subfigure}
    \hfill
    \begin{subfigure}[b]{.24\textwidth}
    \centering
        \includegraphics[width=\textwidth]{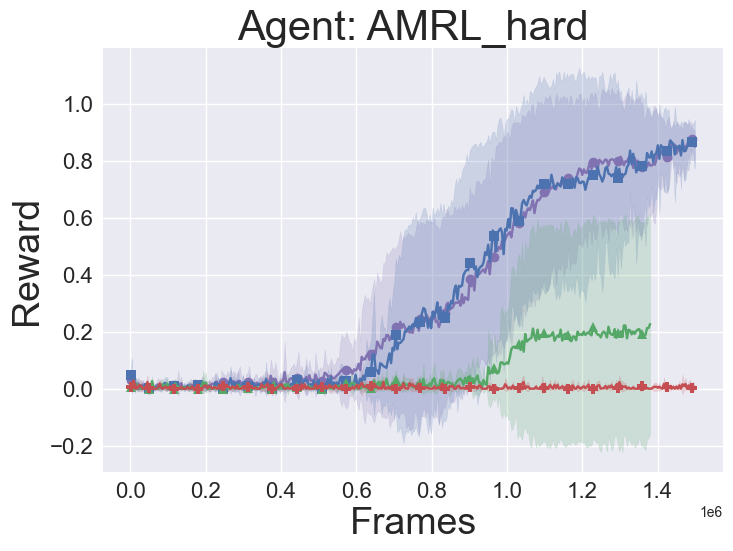}
    \end{subfigure}
    \hfill
    \begin{subfigure}[b]{.24\textwidth}
    \centering
        \includegraphics[width=\textwidth]{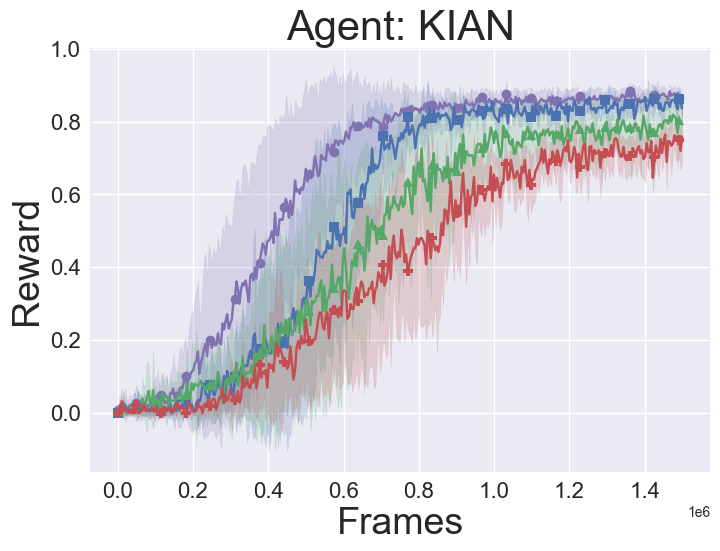}
    \end{subfigure}
    \hfill
    \begin{subfigure}[b]{.24\textwidth}
    \centering
        \includegraphics[width=\textwidth]{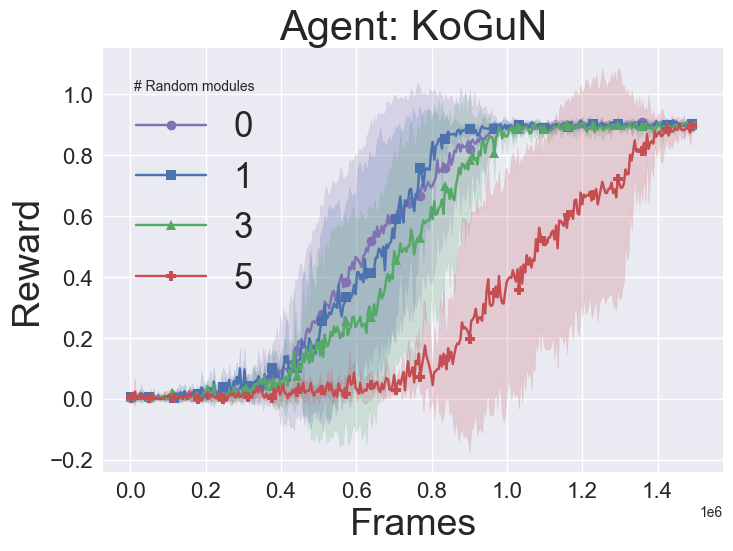}
    \end{subfigure}
    \caption{The reward dynamics during training on MiniGrid-LavaCrossing-S9N1 (training for $1.5 \times 10^6$ frames). The set of original modules is modified by adding 1, 3, and 5 additional modules outputting uniformly random actions. AMRL demonstrates consistent robustness across varying numbers of random modules, while AMRL$_{hard}$, KIAN, and KoGuN exhibit higher sensitivity, with performance and sample efficiency degrading significantly as the number of random modules increases.}
    \label{fig:noisy}
\end{figure*}

\subsection{Safer Training and Deployment}

A strong motivation for utilizing heterogeneous knowledge in RL is to improve safety during training and deployment. All agents incorporating heterogeneous knowledge achieve significantly higher performance than PPO on the safety critical Lava Crossing environments, which directly corresponds to fewer unsafe actions.
We record unsafe actions, such as crashing into a wall or coming in contact with the lava, during training on the LavaCrossing S9N1 environment; we plot the percentage of unsafe actions performed by each agent in Figure \ref{fig:safety}. AMRL performs the smallest amount of unsafe actions averaged across training runs.

Beyond the above safety improvements achieved by incorporating heterogeneous knowledge, module prioritization can be deployed with AMRL for further gains. Given that a module is known to produce safe and trustworthy actions only, such as the rule module, module prioritization can be deployed to shortcut the decision making process and only act according to the safe module in critical states. This will reduce unsafe actions to $0\%$ if the module is perfectly accurate.

\subsection{Improving Interpretability through Modularity}
The modularity of AMRL combined with heterogeneous knowledge significantly improves interpretability. Heterogeneous knowledge sources, such as rules and skills, are inherently more interpretable than a neural network-based policy trained on the fly. Assuming interpretable modules, an analysis of the selector's outputs is sufficient to gain an insight into performed actions.
Additionally, we note that the number of modules is likely to be smaller than that of the primitive actions, reducing the complexity of an analysis. Plotting the selector's weights during an evaluation episode in DoorKey (Figure \ref{fig:selector:weights}) shows the expected order of first relying on the Unlock skill and then the dynamic module. Additional results for other environments and agents can be found in Appendix \ref{app:selector_weights}.

\subsection{Module Informativeness}
We investigate the effect of module informativeness,  i.e. how informative it is for the task at hand as discussed in Subsection~\ref{ssec:knowledge}, further by training agents accessing modules with high, medium, and low informativeness (details in Appendix~\ref{app:exp_details}). The results are reported in Table~\ref{tab:informativeness}. We find that while no agent with low module informativeness is able to solve the DoorKey 8x8 environment after 300k frames, AMRL with soft selection shows a slight advantage for medium informativeness and again significantly outperforms for high informativeness. On the Empty 16x16 environment low, medium and high knowledge has a positive impact on the agent's efficiency, but again AMRL benefits most. The larger standard deviations are due to the fact that we have stopped training earlier (200k frames) than in Table~\ref{tab:perf_envgen}. The training plots for these results can be found in Appendix~\ref{app:additional_results}.

\subsection{Robustness against Noisy Modules}
Having verified the expected effects of differences in module informativeness for a fixed number of modules, we now test the robustness of the selector against noisy modules. Specifically, we vary the density of the informative knowledge contained in all modules by adding in one, three, and five random modules, i.e. modules outputting equal action preferences for all actions. In Table \ref{tab:noisy} we report the achieved performance in terms of reward and Figure \ref{fig:noisy} shows the training runs on the LavaCrossing S9N1 environment. The training runs for the other environments can be found in Appendix \ref{app:noisy}. 

AMRL generally performs consistently well across environments, showing minimal degradation even as the number of random modules increases. AMRL$_{hard}$ shows poor performance under random modules in most environments. KIAN displays moderate robustness. Its performance degrades consistently as randomness increases. There is significant variability in performance as indicated by the larger standard deviations in some environments (e.g., Door Key 8x8). KoGuN exhibits robustness similar to AMRL, showing consistently strong performance across most environments. Additionally, Figure \ref{fig:noisy} shows that in contrast to AMRL which performs consistently, KoGuN performs well under moderate randomness, comparable to AMRL, but its sensitivity to higher levels of randomness (5 modules) reveals that it is more affected by high levels of randomness. 

Overall, AMRL is the most robust and resilient method when training in stochastic environments with random modules. It learns consistently and reaches near-optimal rewards regardless of the number of random modules. KoGuN has slightly worse performance but exhibits more variability, especially when the number of random modules is high. KIAN is moderately robust but less reliable, and AMRL$_{hard}$ is confirmed to be sensitive to randomness, almost entirely failing when random modules are introduced.

Beyond the demonstration of robustness, the results presented in Table \ref{tab:noisy} and Figure \ref{fig:noisy} can be used for assessing the scaling properties of the architecture as the number of modules increases from four modules to nine. AMRL scales generally well, while other methods struggle to identify and utilize the relevant modules as their number increases.
\subsection{Limitations and Potential Extensions}
AMRL suffers from the following limitations. Firstly, it is worth noting that the complexity of the selector's action space grows with the number of modules; however, in practical settings, we expect that the number of modules will be limited. Secondly, naturally, when using existing knowledge its relevance and quality is crucial and AMRL is not an exception. However, we emphasize that heterogeneous knowledge might help in developing solutions that are possibly more robust in situations characterized, for example, by the presence of modules trained with noisy data or based on inconsistent rules.

The release of new Large Language Models (LLM) have inspired a large body of work leveraging LLMs within RL agents, among others, to provide intrinsic rewards \citep{klissarov2023motif}, guide exploration \citep{du_guiding_2023}, perform planning \citep{ahn_as_nodate, huang_adarl_2022}, and satisfy safety constraints \citep{yang_safe_2021}. LLMs are indeed a rich knowledge base of human preferences and as such are compatible with our proposed approach. Future work could investigate incorporating modules based on LLMs, for example, to provide high level planning abilities and additional interpretability of the system. In fact, AMRL is designed to be an open and flexible architecture, able to support learning mechanisms that are potentially not available yet.
\section{Conclusion}

In this paper we have discussed the design, implementation, and evaluation of Augmented Modular Reinforcement Learning (AMRL), a solution that is able to process and integrate multiple types of information (heterogeneous knowledge), such as rules, trajectory datasets, and skills for more efficient and effective decision-making. We have formally characterized heterogeneous knowledge and investigated several sources of information heterogeneity.
We have evaluated the proposed mechanisms on several Minigrid environments and benchmark them against KIAN, KoGuN, and PPO.
Our proposed framework uses a selector to choose exactly one or combine several heterogeneous modules and is able to seamlessly incorporate different types of knowledge. This selector is agnostic with respect to the knowledge representation itself. 

While heterogeneous knowledge benefits all agents accessing it, our results show performance and efficiency improvements achieved by AMRL in comparison to the benchmark methods.
We have shown its flexibility and its potential, especially with respect to applications that require safe training and exploration. We have also demonstrated how AMRL improves the overall interpretability of a learning system through modularity and heterogeneous knowledge. Finally, we experimentally demonstrated AMRL's robustness in presence of noisy modules. 

\bibliography{references_editable}
\bibliographystyle{arxiv}

\include{arxiv_supplementary}

\end{document}

%% file: preamble.tex
\newcommand{\mm}[1]{\textcolor{red}{""MM: #1""}}
\newcommand{\lw}[1]{\textcolor{teal}{""LW: #1""}}

\newcommand{\cut}[1]{}
\newcommand{\actionspace}{\mathcal{A}}
\newcommand{\statespace}{\mathcal{S}}

%% file: arxiv_supplementary.tex
\appendix

\section{Policy Gradients for AMRL} 
\label{app:gradients}
\subsection{Notation}
Given a policy $\pi$ such that $\pi(a,s)=Pr(a|s)$ is the probability of taking action $a$ in state $s$ when following policy $\pi$ and assuming we have an finite set of actions $\actionspace=\{a_i\}_{i=1}^{|\actionspace|}$, we use $\pi(s)$ to denote the vector with ith component being  $\pi(a_i|s)$. Additionally, given a list of modules $\mathbf{M}$ with corresponding policies $\{\pi_M\}_{M\in\mathbf{M}}$ we use $\boldsymbol{\Pi}_{\mathbf{M}}(a|s)$ to denote the column vector with ith entry $\pi_M(a|s)$ where $M$ is the ith module in $\mathbf{M}$.

Furthermore, denote the selectors policy parameterized by $\phi$ as $\pi^{\phi}_{selector}(\cdot|s)$ and the policy of the dynamic module as $\pi^{\theta}_{dyn}(\cdot|s)$. Let the set $\mathbf{M}$ contain all modules of which one is the dynamic module.

\subsection{Policy Gradients}
The vanilla Policy Gradient of a policy $\pi^{\theta}$ is given by:
$$\nabla_{\theta} J\left(\pi^{\theta}\right)=\underset{\tau \sim \pi^{\theta}}{\mathrm{E}}\left[\sum_{t=0}^T \nabla_{\theta} \log \pi^{\theta}\left(a_t \mid s_t\right) A^{\pi^{\theta}}\left(s_t, a_t\right)\right],
$$ where $J$ is the un-discounted finite time horizon reward to-go.
Furthermore, the update rule used in PPO \citep{schulman2017proximal} is given by
\begin{align}
\theta_{k+1}=&\arg \max _\theta \frac{1}{\left|\mathcal{D}_k\right| T} \sum_{\tau \in \mathcal{D}_k} \sum_{t=0}^T \min \left(\frac{\pi_\theta\left(a_t \mid s_t\right)}{\pi_{\theta_k}\left(a_t \mid s_t\right)} A^{\pi_{\theta_k}}\left(s_t, a_t\right), \quad g\left(\epsilon, A^{\pi_{\theta_k}}\left(s_t, a_t\right)\right)\right).
\end{align}
While the below analysis is focused on the vanilla policy gradient, it justifies the use of PPO to train the selector.

\subsection{Soft Selection}
Using the specified notation we can write the AMRL policy with soft selection as:
$$
\pi^{\phi,\theta}(a|s) = \pi^{\phi}_{selector}(s)^T \boldsymbol{\Pi}_{\mathbf{M}}(a|s)
$$
where $\mathbf{M}$ is the set of modules, of which the only module parameterized is the dynamic module $M_{dyn}$ with policy $\pi_{dyn}^{\theta}$.
Alternatively, we can write:
$$\pi^{\phi,\theta}(a|s) = \pi_S^{\phi}(M_{dyn}|s)\pi_{M_{dyn}}^{\theta}(a|s) + \sum_{M\in \mathbf{M}\setminus M_{dyn}} \pi_S^{\phi}(M|s)\pi_{M}(a|s),$$
where we have written out the inner product and split out the term corresponding to the dynamic module.

We can compute the policy gradients with respect to each of the parameter sets $\phi$ and $\theta$ as follows. The gradient w.r.t. the dynamic module parameters $\theta$ can be computed as:
\begin{align}
    \nabla_{\theta}J(\pi^{\phi,\theta}) 
        &= \underset{\tau \sim \pi_\theta}{\mathrm{E}}\left[\sum_{t=0}^T \nabla_\theta \log \pi^{\phi,\theta}\left(a_t \mid s_t\right) A^{\pi^{\phi,\theta}}\left(s_t, a_t\right)\right]\\
\end{align}
We can then derive:
\begin{align}
    \nabla_\theta \log \pi^{\phi,\theta}\left(a_t \mid s_t\right) 
        &= \frac{1}{\pi^{\phi,\theta}\left(a_t \mid s_t\right)} \nabla_{\theta} \pi^{\phi,\theta}\left(a_t \mid s_t\right)\\
        &= \frac{1}{\pi^{\phi,\theta}\left(a_t \mid s_t\right)} \pi_S^{\phi}(M_{dyn}|s_t) \nabla_{\theta} \pi_{dyn}^{\theta}(a_t|s_t).
\end{align}
Similarly, we compute the policy gradient w.r.t. the selector's parameters $\phi$, as follows:
\begin{align}
    \nabla_\phi \log \pi^{\phi,\theta}\left(a_t \mid s_t\right) 
        &= \frac{1}{\pi^{\phi,\theta}\left(a_t \mid s_t\right)} \nabla_{\phi} \pi^{\phi,\theta}\left(a_t \mid s_t\right)\\
        &= \frac{1}{\pi^{\phi,\theta}\left(a_t \mid s_t\right)} \nabla_{\phi} \boldsymbol{\pi^{\phi}_{selector}}(s)^T \boldsymbol{\Pi}_{\mathbf{M}}(a|s)\\
        &= \frac{1}{\pi^{\phi,\theta}\left(a_t \mid s_t\right)} 
        \Big( \nabla_{\phi}\pi_S^{\phi}(M_{dyn}|s_t)\pi_{dyn}^{\theta}(a_t|s_t) + \sum_{M\in \mathbf{M}}\nabla_{\phi}\pi_S^{\phi}(M|s_t)\pi_{M}(a_t|s_t)\Big)\\
\end{align}

\subsection{Hard Selection}
In the case of hard selection we use the Gumbel Softmax trick to perform differentiable selection. In particular, we want to sample from a categorical distribution where the probabilities are given by the selector as $\boldsymbol{\pi_S^{\phi}}(s)$. In other words, we select module $M$ with probability equal to $\pi_{selector}^{\phi}(M|s)$. We then obtain a sample vector $\mathbf{y}$ of dimensions $|\actionspace_{selector}|$ by setting:
$$
\mathbf{y}_i=\frac{\exp \left(\left(\log \left(\pi_{selector}^{\phi}(M_i|s)\right)+g_i\right) / \tau\right)}{\sum_{j=1}^k \exp \left(\left(\log \left(\pi_{selector}^{\phi}(M_j|s)\right)+g_j\right) / \tau\right)} \quad \text { for } i=1, \ldots, k ,
$$
\noindent where $g_1 \ldots g_k$ are i.i.d samples drawn from $\operatorname{Gumbel}(0,1)$, which can be easily sampled using an inverse transform. Note that the softmax function is used as a continuous differentiable approximation. As the temperature $\tau$ tends to $0$, the samples $\mathbf{y}$ tend to the samples of the true categorical distribution \cite{jang2017categorical}.

\section{Environments}
\label{app:envs}
In all the below environments the action space consists of 7 discrete actions. The observation size is 7x7.\\

\noindent\textit{DoorKey.}
In this environment the agent is required to first pick up a key, then unlock a door and reach a goal square. We us the 6x6 and 8x8 versions of this environment for evaluation and the 5x5 version for some of the knowledge modules.\\

\noindent\textit{LavaCrossing.}
In this environment the agent needs to navigate through lava streams (with one square openings) to reach a goal square. Entering a lava stream immediately ends the episode. We use the 9x9 version with 1 and 2 lava streams flowing across the grid.\\

\noindent\textit{Empty.}
In this environment the agent must reach a goal square, there are no additional challenges. We use the 16x16 version for evaluation and the 8x8 and 5x5 as well as 5x5 random version for some of the knowledge modules.

\section{MiniGrid Rules}
\label{app:rules}
The rules used for evaluation on the MiniGrid environments are the following:
\begin{itemize}
    \item If there is a key inside the state, then pick it up if possible or else move towards it.
    \item If there is a ball inside the state, then pick it up if possible or else move towards it.
    \item If there is a door inside the state, then open it if possible or else move towards it.
    \item If there is a goal inside the state, then move towards it.
    \item If there is Lava in front of you, do not move forward.
    \item If there is a wall in front of you, do not move forward.
\end{itemize}

\section{Additional Implementation Details}
\label{appendix:reproducibility}
For all agents below the PPO hyperparameters are set to the default values provided in the \verb+rl-starter-file+ repository\\ (\href{https://github.com/lcswillems/rl-starter-files}{https://github.com/lcswillems/rl-starter-files}).
\\
\noindent\textit{PPO} We use the network architecture of actor and critic network as implemented in the \verb+rl-starter-files+ repository.
In particular:
\begin{itemize}
    \item \textit{Embedding}: A CNN with 3 convolutional layers produces an embedding of the image observations. Both the actor and the critic rely on these image embeddings.
    \item \textit{Actor}: The actor network is a 2 layer fully connected neural network with layer width 64.
    \item \textit{Critic}: The critic network is a 2 layer fully connected neural network with layer width 64.
    
\end{itemize}

\noindent\textit{AMRL} In the following, we summarize the key design and implementation choices of our algorithm:
\begin{itemize}
    \item \textit{Embedding}: Same as in PPO. The Embedding network is shared by the selector, the dynamic RL module, and the critic network.
    \item \textit{Selector}: The selector network is a 2 layer fully connected neural network with layer width 64. It outputs a weight for each module. The softmax function is applied to the weights to map them to probabilities. To sample a module in the hard selection mechanism we use the Gumbel softmax distribution.
    \item \textit{Critic}: Same as in PPO.
\end{itemize}

\noindent\textit{KIAN} We rely on the original implementation provided by the authors \href{https://github.com/Pascalson/KGRL}{https://github.com/Pascalson/KGRL}.
\begin{itemize}
    \item \textit{Embedding}: A CNN with 3 convolutional layers produces an embedding of the image observations. The inner actor, the key network, and the critic rely on these image embeddings.
    \item \textit{Actor}: The actor consists of an internal actor, the query network, and a key network. All 3 share an initial fully connected base layer of width 64, which is followed by an additional fully connected layer mapping from 64 features to the desired dimension (embedding size of 8 for the query and key network, and $|\actionspace|$ for the internal actor). The keys of modules are learned by the nn.Embedding layer.
    \item \textit{Critic}: Same as in PPO.
\end{itemize}

\noindent\textit{KoGuN} The following summarizes the key design choices of the KoGuN implementation.
\begin{itemize}
    \item \textit{Embedding}: Same as for PPO. Both the actor, and the critic rely on these image embeddings.
    \item \textit{Actor}: The actor network is a 2 layer fully connected neural network with layer width 64. It takes as inputs the output of the embedding network concatenated with the action preferences averaged across modules.
    \item \textit{Critic}: Same as in PPO.
\end{itemize}

\noindent\textit{Modules.} All modules are implemented as described in the body of the main paper.\\

\noindent\textit{Compute Resources.} The experiments were run on a CPU. No large amount of memory is required.

\section{Computational Cost}
An analysis of the computational cost of AMRL and baselines can be found in Table \ref{tbl:computational_cost}.
\label{app:cost}
\begin{table}[htbp]
\caption{Analysis of the computational cost. The number of trainable parameters includes the trainable dynamic module for AMRL and the inner actor for KIAN. All knowledge based architectures are comparable and rely on the skill, retrieval and rule modules. The retrieval module searches for 4 neighbors in a dataset of 3040 transitions. The faster inference times of PPO are caused by not calling the knowledge modules.}
    \centering
    \begin{tabular}{ccc}
        \hline
        \textbf{Method} & \textbf{\# Trainable params} & \textbf{Total inference time (s)} \\
        \hline
        PPO & 19384 & $3.4 \times 10^{-5}$\\
        AMRL & 19644 & $1.5 \times 10^{-4}$\\
        KoGuN & 19832 & $1.5 \times 10^{-4}$ \\
        KIAN & 20448 & $1.6 \times 10^{-4}$\\
        \hline
    \end{tabular}
    \label{tbl:computational_cost}
\end{table}

\section{Additional Experimental Details}
\label{app:exp_details}

\noindent\textit{DoorKey 8x8.} The knowledge configurations of different levels used for the DoorKey 8x8 environment is as follows:
\begin{itemize}
    \item \textit{Low:} Skill - Empty 5x5, Retrieval - Empty 5x5, Rules - No rules
    \item \textit{Medium}: Skill - DoorKey 5x5, Retrieval - DoorKey 5x5, Rules - All rules
    \item \textit{High:} Skill - DoorKey 6x6, Retrieval - DoorKey 6x6, Rules - All rules
\end{itemize}

\noindent\textit{Empty 16$\times$16.} On Empty the knowledge for different informativeness levels is:
\begin{itemize}
    \item \textit{Low:} Skill - Empty 5x5, Retrieval - Empty 5x5, Rules - No rules
    \item \textit{Medium:} Skill - Empty 8x8, Retrieval - Empty 8x8, Rules - All rules
    \item \textit{High:} Skill - Empty 16x16, Retrieval - Empty 16x16, Rules - All rules
\end{itemize}

\section{Additional Results}
\label{app:additional_results}

\subsection{Reward Plots}
\begin{figure*}[htbp]
    \centering
    \begin{subfigure}[b]{.24\textwidth}
        \centering
        \includegraphics[width=\textwidth]{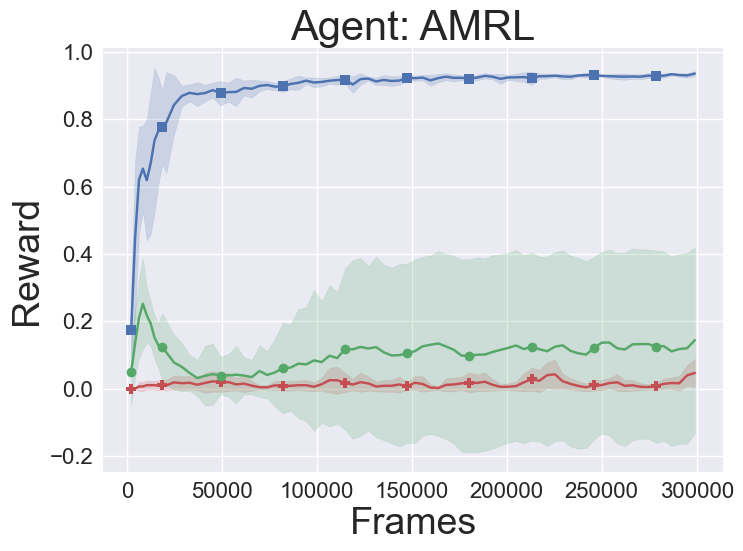}
    \end{subfigure}
    \hfill
    \begin{subfigure}[b]{.24\textwidth}
    \centering
        \includegraphics[width=\textwidth]{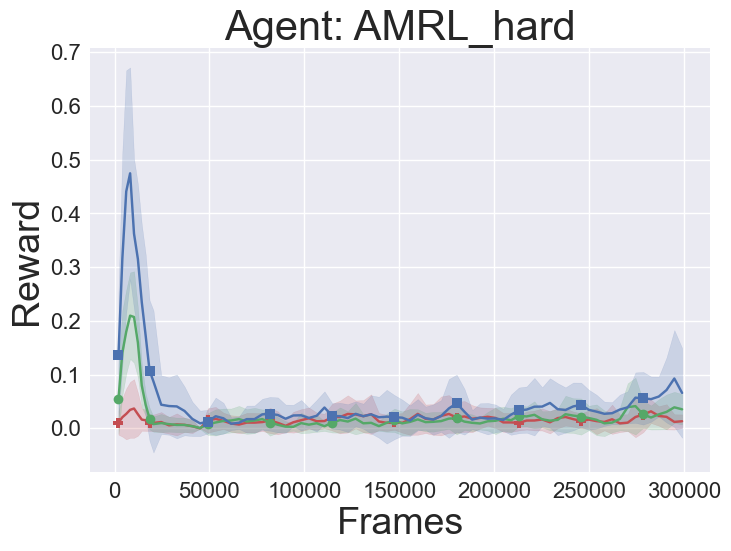}
    \end{subfigure}
    \hfill
    \begin{subfigure}[b]{.24\textwidth}
    \centering
        \includegraphics[width=\textwidth]{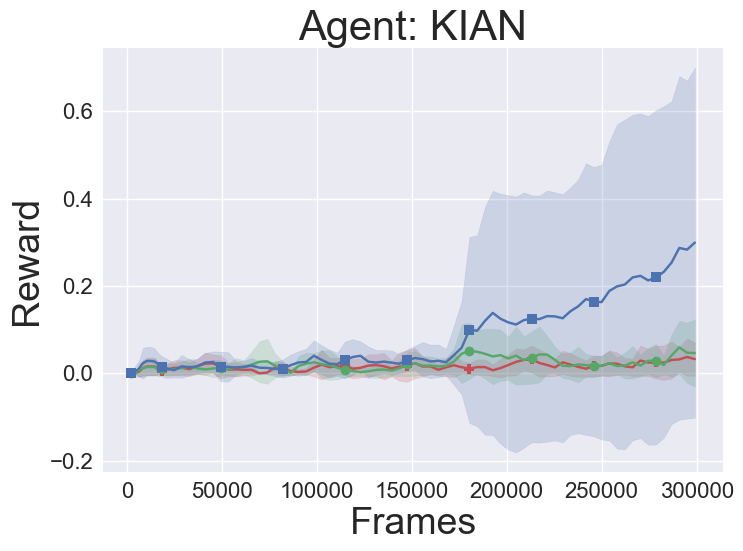}
    \end{subfigure}
    \hfill
    \begin{subfigure}[b]{.24\textwidth}
    \centering
        \includegraphics[width=\textwidth]{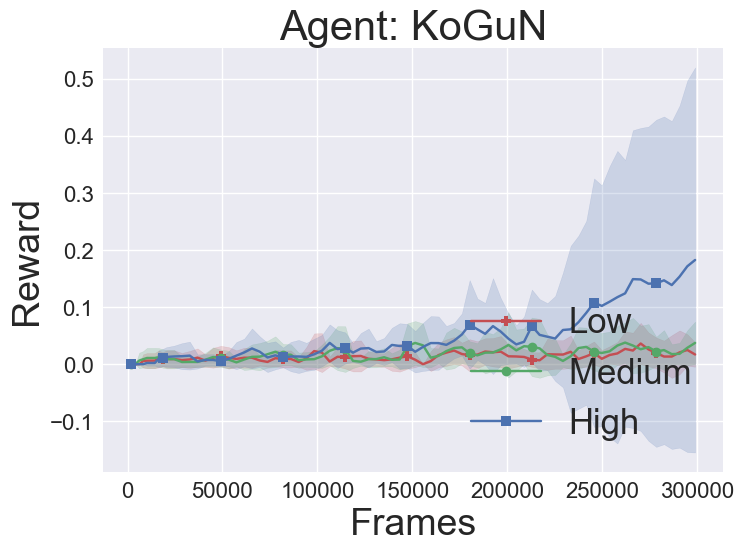}
    \end{subfigure}
    \caption{Average training run performance on DoorKey 8x8 with different levels of knowledge.}
    \label{fig:module_info_doorkey}
\end{figure*}

\begin{figure*}[htbp]
    \centering
    \begin{subfigure}[b]{.24\textwidth}
        \centering
        \includegraphics[width=\textwidth]{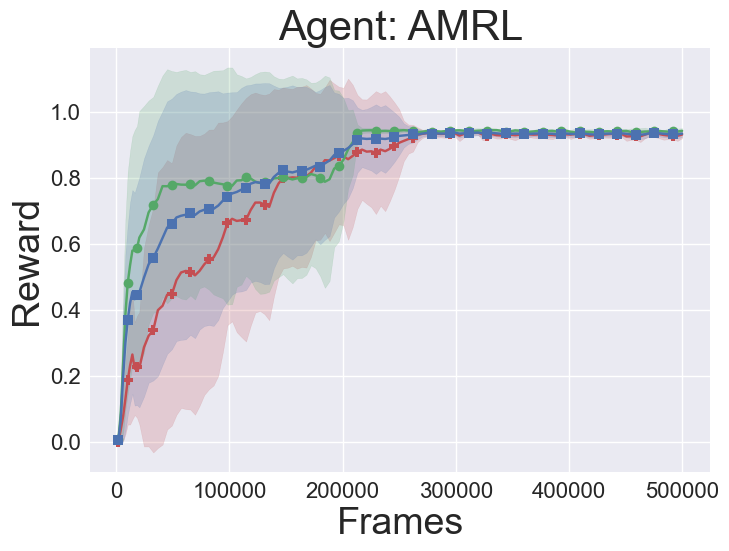}
    \end{subfigure}
    \hfill
    \begin{subfigure}[b]{.24\textwidth}
    \centering
        \includegraphics[width=\textwidth]{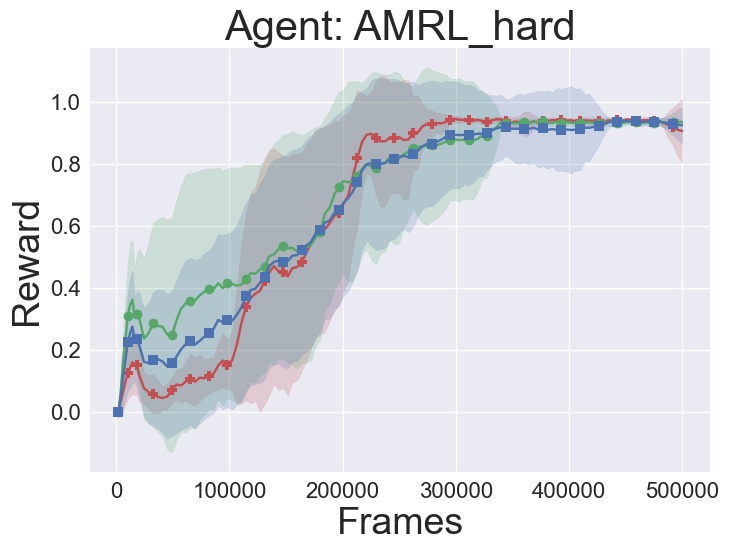}
    \end{subfigure}
    \hfill
    \begin{subfigure}[b]{.24\textwidth}
    \centering
        \includegraphics[width=\textwidth]{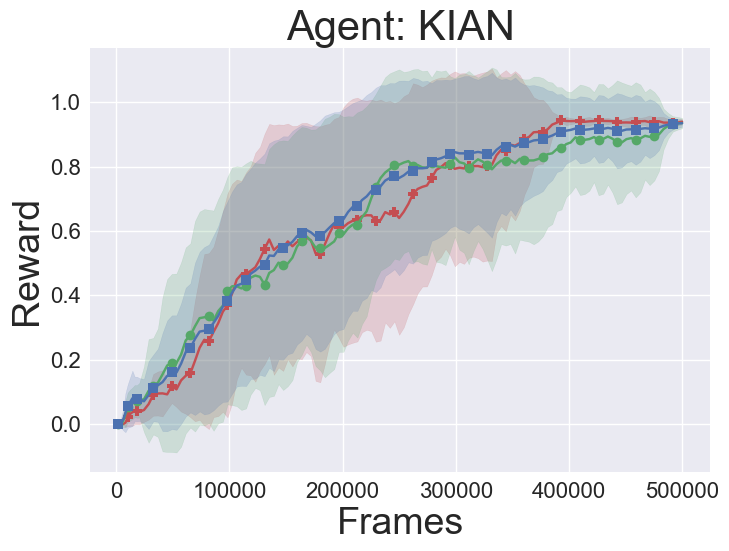}
    \end{subfigure}
    \hfill
    \begin{subfigure}[b]{.24\textwidth}
    \centering
        \includegraphics[width=\textwidth]{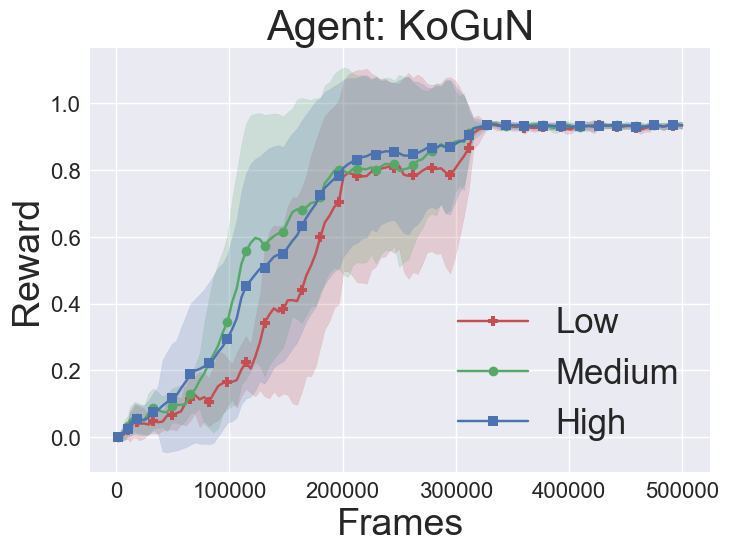}
    \end{subfigure}
    \caption{Average training run performance on Empty 16x16 with different of levels of knowledge. After 500k all agents have converged to roughly equal performance. We shorten the training runs and instead reported the performance after 200k frames of training.}
    \label{fig:module_info_empty}
\end{figure*}

For completeness we present the training runs for the module informativeness experiment in Figures~\ref{fig:module_info_doorkey} and \ref{fig:module_info_empty}.

\subsection{Analyzing Learned Policies}
\label{app:selector_weights}
\begin{figure*}
    \centering
    \includegraphics[width=0.8\textwidth]{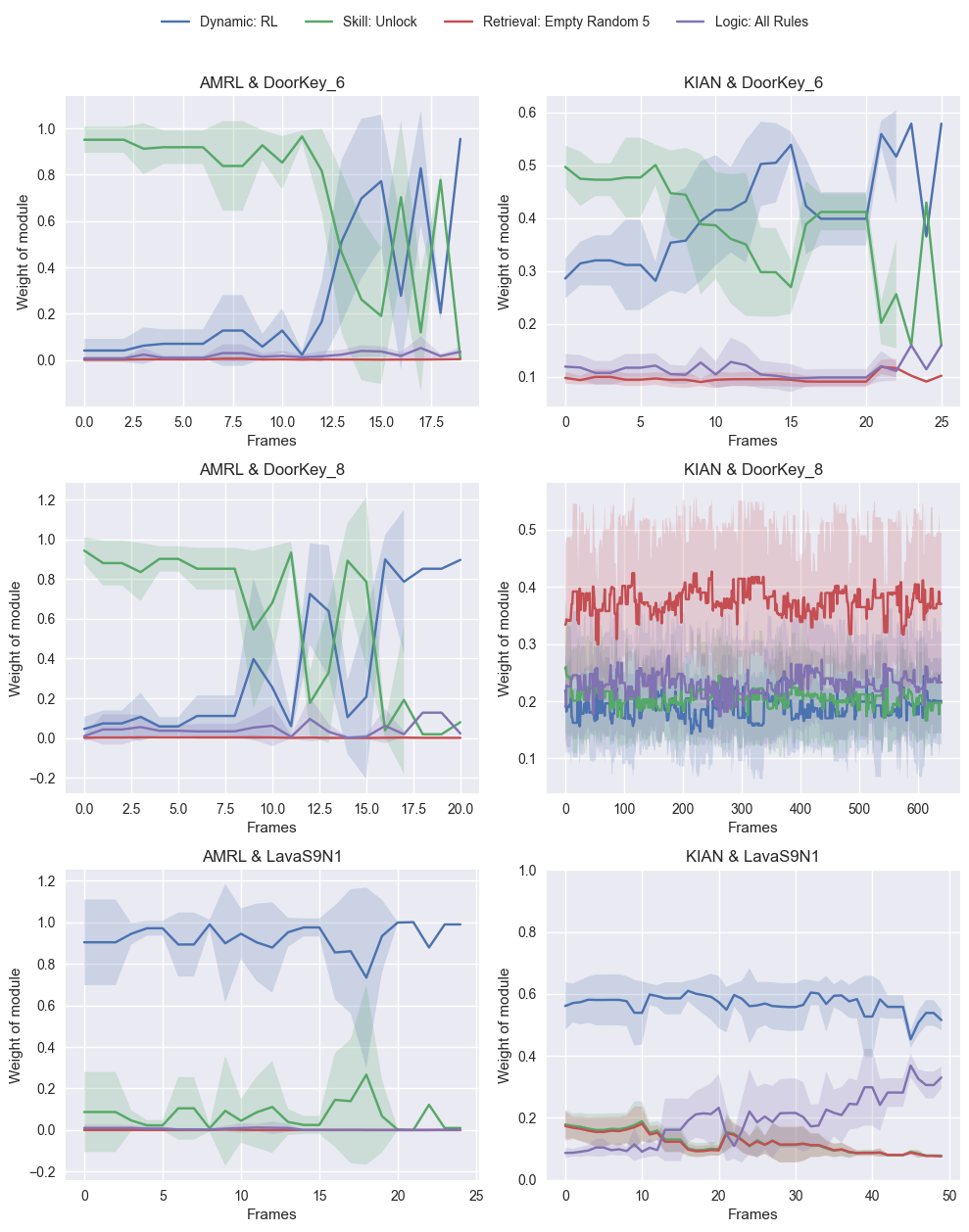}
    \caption{Weightings of modules in the final policy. Note that in contrast to AMRL, KIAN has not converged on the DoorKey $8\times 8$ environment and seems to be stuck because it puts too much weight on the retrieval module instead of the more relevant skill module. In the LavaCrossing S9N1 environment AMRL has converged to almost fully rely on the dynamic RL module in contrast to KIAN which has slightly more balanced weights.}
    \label{fig:weights_all}
\end{figure*}

To further investigate the learned policies and showcase improved interpretability,
we plot the weights of the Selector in AMRL and the weights of the attention mechanism in KIAN to compare the learned policies in a range of selected environments. To obtain the weights, the agents are evaluated in the specified environments, weights are recorded for one episode (same configuration) from start to finish for all training seeds and we report mean weights together with $\pm2$ standard deviation confidence intervals. Note that weights are not comparable across different start positions and environment instances which is why we reported weights for the same episode. The extracted weights are shown in Figure \ref{fig:weights_all}.

\subsection{Gumbel Temperature Ablation}
\label{app:gumbel}

We perform an ablation on the temperature parameter of the Gumbel-Softmax distribution used for the hard selection mechanism. The results of the ablation study are reported for the following two cases:
\begin{itemize}
    \item The samples are discretized and the smooth approximation is used for gradients as in the experiments reported in the paper.
    \item The samples are not discretized.
\end{itemize}
\begin{figure*}
    \centering
    \begin{subfigure}[b]{.47\textwidth}
        \centering
        \includegraphics[width=\textwidth]{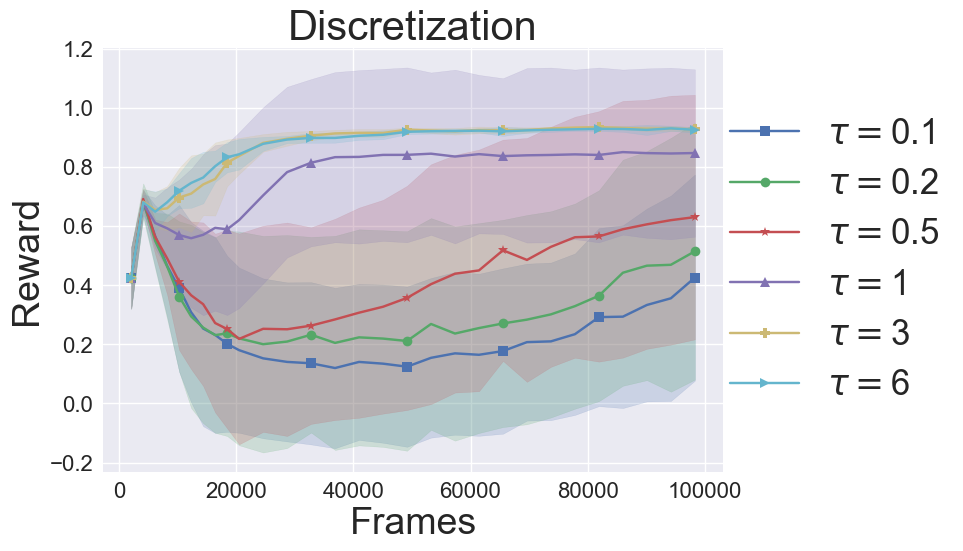}
        \caption{}
    \end{subfigure}
    \hfill
    \begin{subfigure}[b]{.47\textwidth}
    \centering
        \includegraphics[width=\textwidth]{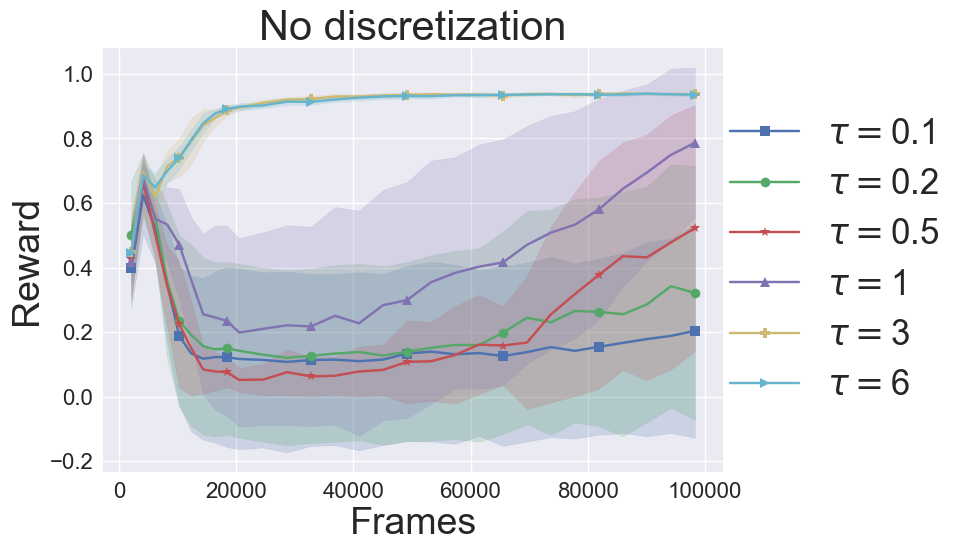}
        \caption{}
 
    \end{subfigure}
    \caption{(a) Ablation on the temperature with discretization in the forward pass. (b) Ablation on the temperature without discretization in the forward pass.}
    \label{fig:gumbel}
\end{figure*}
For both cases we train for 100k frames on the DoorKey 6$\times$6 environment and plot the mean rewards with confidence intervals across 10 random training seeds. The results are shown in Figures $\ref{fig:gumbel}$.(a) and $\ref{fig:gumbel}$.(b) for the discretized and not discretized cases respectively. We find that larger $\tau$ around $3$ performs best in both cases. Additionally, the performance drops as $\tau$ decreases to $0$ and the samples from the categorical distribution become more and more discrete regardless of whether we discretize samples to a one-hot vector in the forward pass.

\subsection{Presence of Random Modules}
\label{app:noisy}

\begin{figure*}[htbp]
    \centering
    \begin{subfigure}[b]{.24\textwidth}
        \centering
        \includegraphics[width=\textwidth]{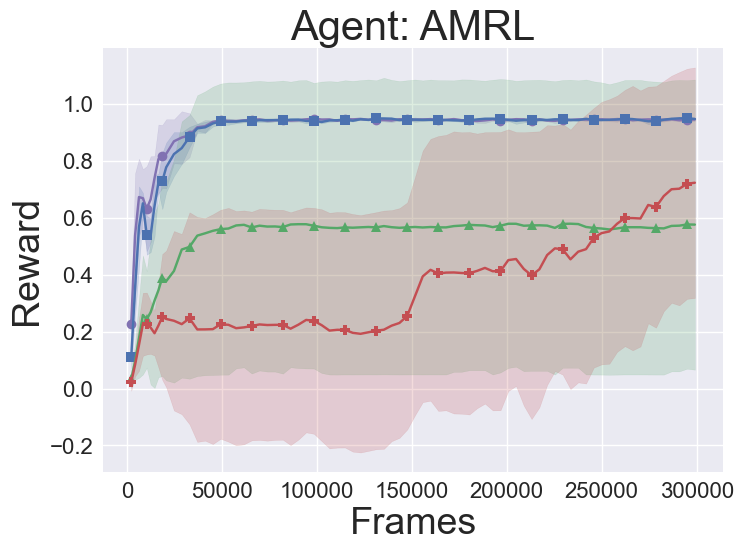}
    \end{subfigure}
    \hfill
    \begin{subfigure}[b]{.24\textwidth}
    \centering
        \includegraphics[width=\textwidth]{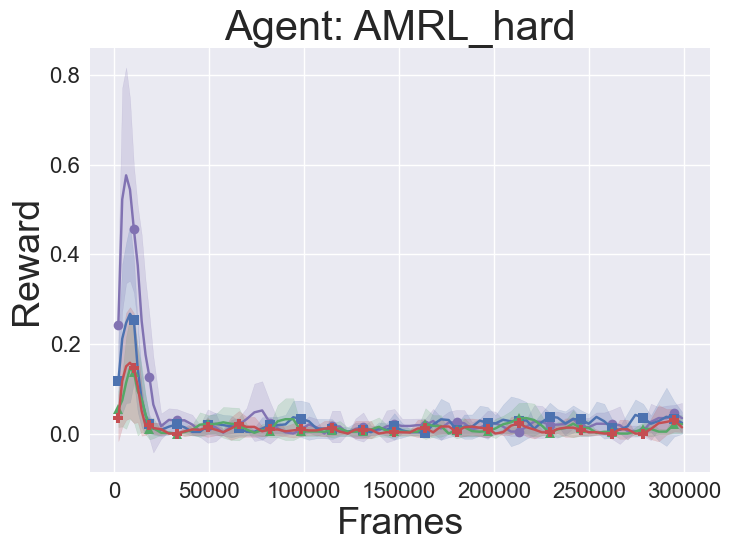}
    \end{subfigure}
    \hfill
    \begin{subfigure}[b]{.24\textwidth}
    \centering
        \includegraphics[width=\textwidth]{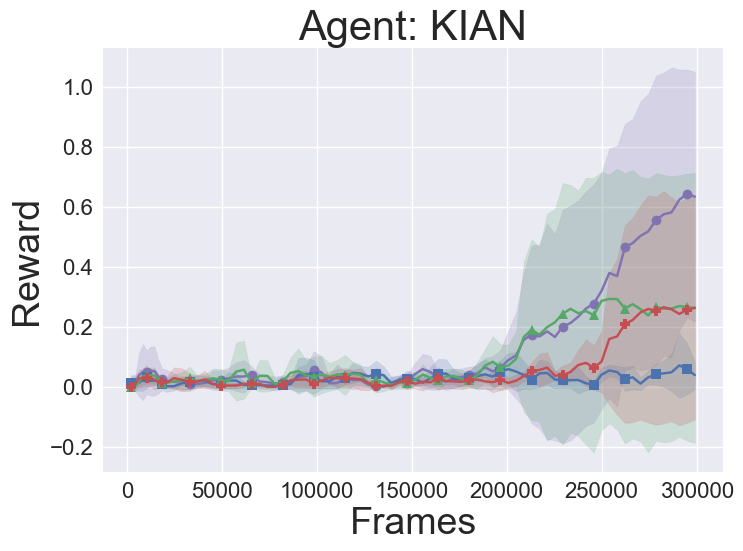}
    \end{subfigure}
    \hfill
    \begin{subfigure}[b]{.24\textwidth}
    \centering
        \includegraphics[width=\textwidth]{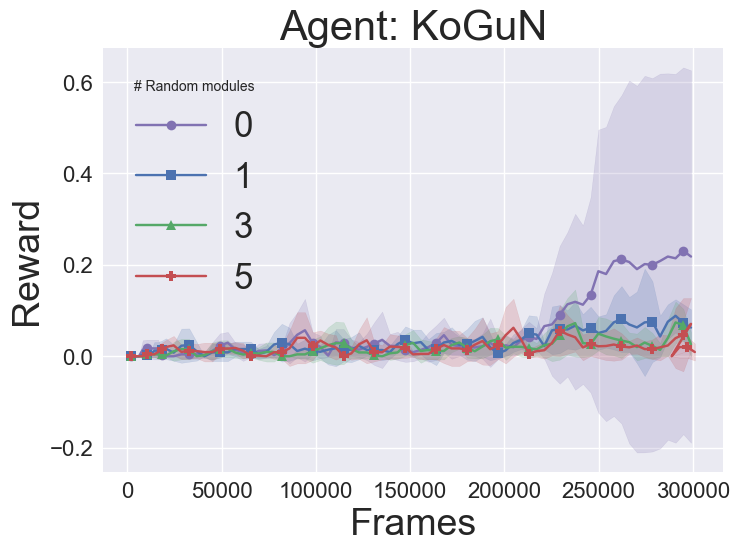}
    \end{subfigure}
    \caption{The achieved reward logged throughout training on Door Key 8x8. The set of original modules is modified by adding 1, 3, and 5 additional modules outputting uniformly random actions. It shows that AMRL can solve the environment with random modules present. However, training is slower for 3 and 5 random modules added. The results for other agents show significantly worse performance when random modules are present, but results are less conclusive due to significantly worse learning in the first place.}
    \label{fig:random_door8}
\end{figure*}

\begin{figure*}[htbp]
    \centering
    \begin{subfigure}[b]{.24\textwidth}
        \centering
        \includegraphics[width=\textwidth]{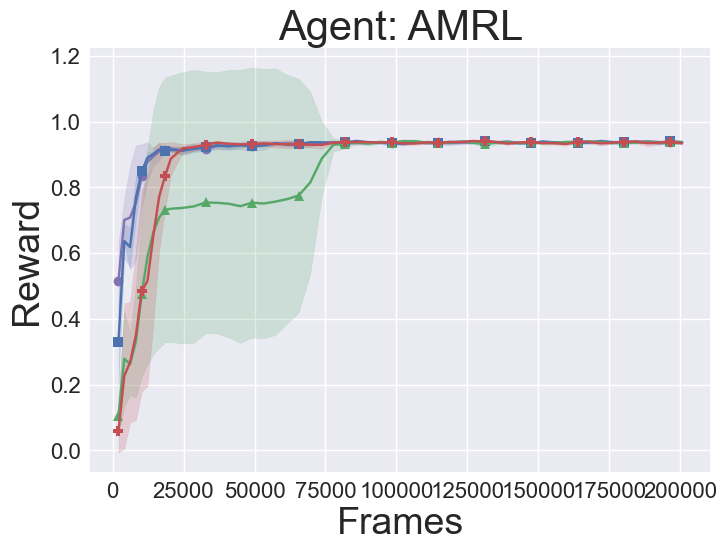}
    \end{subfigure}
    \hfill
    \begin{subfigure}[b]{.24\textwidth}
    \centering
        \includegraphics[width=\textwidth]{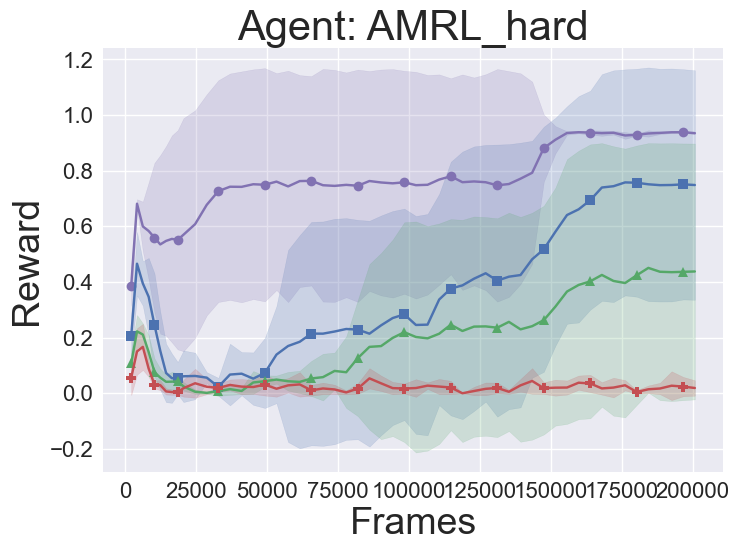}
    \end{subfigure}
    \hfill
    \begin{subfigure}[b]{.24\textwidth}
    \centering
        \includegraphics[width=\textwidth]{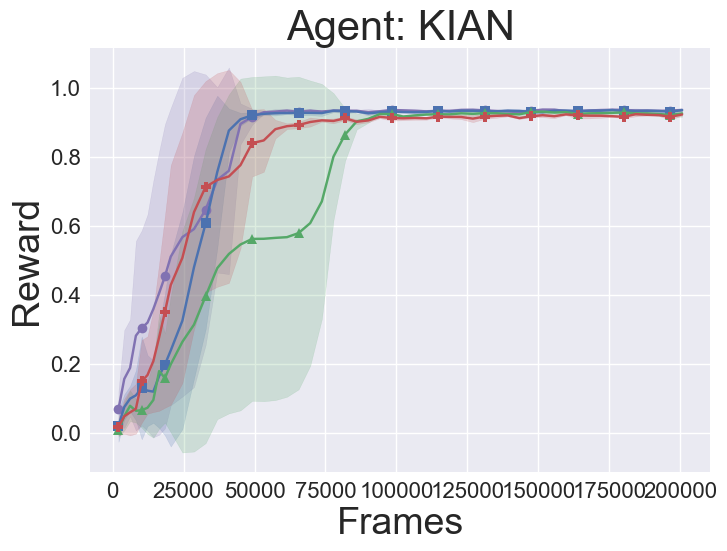}
    \end{subfigure}
    \hfill
    \begin{subfigure}[b]{.24\textwidth}
    \centering
        \includegraphics[width=\textwidth]{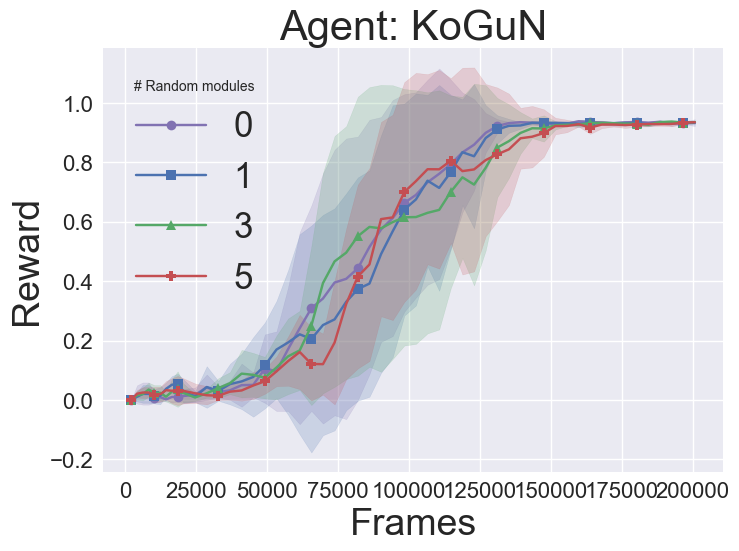}

    \end{subfigure}
    \caption{The achieved reward logged throughout training on Door Key 6x6. The set of original modules is modified by adding 1, 3, and 5 additional modules outputting uniformly random actions.}
    \label{fig:random_door6}
\end{figure*}

\begin{figure*}[htbp]
    \centering
    \begin{subfigure}[b]{.24\textwidth}
        \centering
        \includegraphics[width=\textwidth]{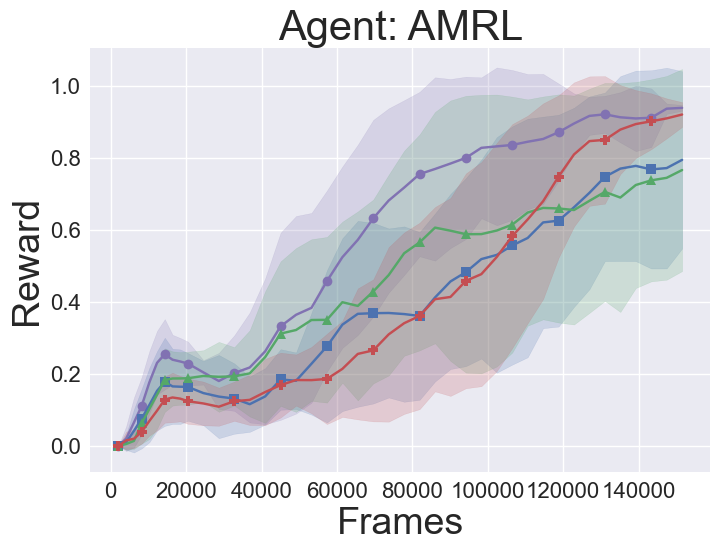}

    \end{subfigure}
    \hfill
    \begin{subfigure}[b]{.24\textwidth}
    \centering
        \includegraphics[width=\textwidth]{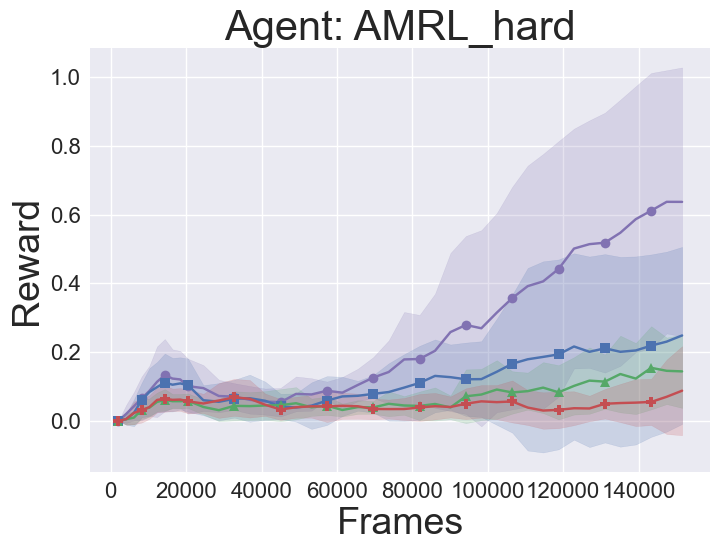}

    \end{subfigure}
    \hfill
    \begin{subfigure}[b]{.24\textwidth}
    \centering
        \includegraphics[width=\textwidth]{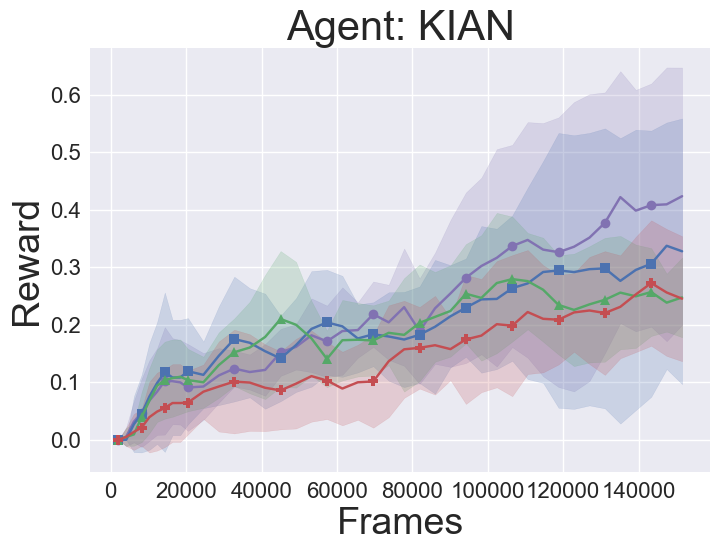}

    \end{subfigure}
    \hfill
    \begin{subfigure}[b]{.24\textwidth}
    \centering
        \includegraphics[width=\textwidth]{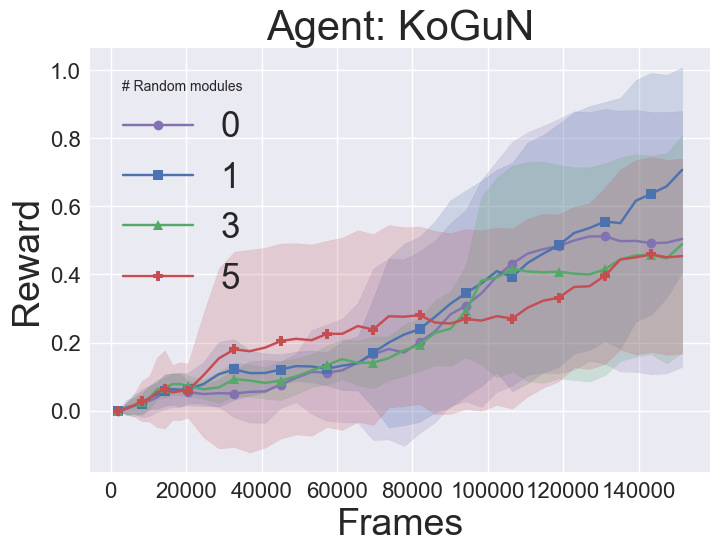}

    \end{subfigure}
    \caption{The achieved reward logged throughout training on MiniGrid-Empty-16x16, training for 150k frames. The set of original modules is modified by adding 1, 3, and 5 additional modules outputting uniformly random actions.}
    \label{fig:random_empty}
\end{figure*}

In Figures \ref{fig:random_door8}, \ref{fig:random_door6}, and \ref{fig:random_empty} we report the training curves on the DoorKey 8x8, DoorKey 6x6, and Empty 16x16 environments, for evaluating robustness in presence of random modules.